\documentclass{article}
\usepackage{iclr2026_conference,times}

\usepackage{amsmath,amsfonts}
\usepackage{amssymb}
\usepackage{enumitem}
\usepackage{graphicx}
\usepackage{booktabs}
\usepackage{algorithm}
\usepackage{algorithmic}
\usepackage{subcaption}
\usepackage{float}
\usepackage{nicefrac}

\usepackage{xcolor}

\newcommand{\blfootnote}[1]{%
 \begingroup\renewcommand\thefootnote{}\footnote{#1}%
 \addtocounter{footnote}{-1}\endgroup}

\title{FARE: Deep Reinforcement Learning for Fair Exposure Constrained Uncertainty-Aware Financial Content Personalization}

\author{%
Arundeep Chinta \\
JPMorganChase \\
Palo Alto, CA, USA \\
\texttt{arundeep.chinta@jpmchase.com} \\
\And
Lucas Vinh Tran \\
JPMorganChase \\
London, UK \\
\texttt{lucas.vinhtran@chase.com} \\
\And
Jay Katukuri \\
JPMorganChase \\
Palo Alto, CA, USA \\
\texttt{jay.katukuri@chase.com} \\
}

\iclrfinalcopy

\usepackage{url}
\usepackage[colorlinks=true,linkcolor=blue,citecolor=blue,urlcolor=blue,breaklinks=true]{hyperref}

\begin{document}
\raggedbottom

\maketitle
\lhead{\small 2nd Workshop on Advances in Financial AI at ICLR 2026}
\blfootnote{Extended version of a paper accepted to the 2nd Workshop on Advances in Financial AI: Towards Agentic and Responsible Systems at ICLR 2026, Rio de Janeiro, Brazil.}

\begin{abstract}

Content personalization systems in financial services must ensure fair exposure across diverse offerings---a requirement driven by contractual obligations and the need to prevent ``rich-get-richer'' dynamics where content with high click-through rate (CTR) dominates while other relevant products receive minimal visibility. Share of Voice (SOV)\footnote{Share-of-Voice (SOV): fraction of top-position exposures per content category over time. Defined formally in Section~\ref{sec:sov}.} constraints, which guarantee each content category a target fraction of top-position exposure, address this by promoting product diversity and balanced user discovery. While re-ranking layers atop CTR models are common in practice, we propose two key novelties: (1) framing SOV-constrained ranking as a deep reinforcement learning problem analogous to constrained trade execution in algorithmic finance, and (2) explicitly incorporating CTR prediction uncertainty ($\sigma$) into the agent's state space and policy design---enabling larger ranking adjustments for high-uncertainty predictions where deviation from CTR-optimal ordering is less costly. We introduce \textbf{FARE (Fair Ranking Executor)}, a modular uncertainty-aware execution layer that translates any black-box CTR model's predictions into SOV-fair rankings without retraining the underlying model. Our uncertainty-weighted proportional control policy (FARE-PC) and learned neural policies (FARE-ES, FARE-PPO) demonstrate that uncertainty-aware approaches can substantially reduce SOV deviation from fairness targets while minimizing engagement loss, with gradient-free evolution strategies outperforming policy gradient methods on synthetic data and the ordering reversing on KuaiRand-Pure.

\end{abstract}

\section{Introduction}
\label{sec:intro}

Personalized re-ranking systems must balance multiple competing objectives: maximizing user engagement while ensuring fair exposure for diverse content \citep{abdollahpouri2020multi}. Traditional approaches optimize purely for predicted engagement metrics such as click-through rate (CTR), but this often results in a \textit{rich-get-richer} phenomenon where high-performing content dominates top positions, leaving other relevant offerings with minimal visibility---and, because that content then accrues the impressions and clicks that train the next model, reinforcing its own advantage \citep{singh2018fairness, morik2020controlling}.

This imbalance arises across many application domains and poses challenges in several recurring ways:
\begin{itemize}
  \item \textbf{Content diversity matters}: Users benefit from exposure to a variety of items (e.g., diverse financial products, news categories, or video genres) rather than repeated promotion of a single high-CTR offering.
  \item \textbf{Business constraints exist}: Operators often have contractual requirements to provide balanced exposure across content categories.
  \item \textbf{Long-term value is important}: Over-exploitation of high-CTR content may reduce discovery of items with higher long-term user value.
\end{itemize}

Existing approaches address only one side of this problem. \textbf{Fairness-aware re-ranking methods} \citep{singh2018fairness, biega2018equity} enforce exposure constraints at the post-ranking stage but treat relevance predictions as deterministic---they do not account for the uncertainty of the underlying model, missing the opportunity to make fairness adjustments where they are cheapest. \textbf{Uncertainty-aware methods} \citep{chapelle2011empirical, chinta2025variance} use prediction uncertainty for exploration---improving the relevance model over time---but do not leverage uncertainty to guide where fairness interventions are applied. \textbf{FARE bridges this gap}: it uses prediction uncertainty ($\sigma^2$) directly to decide \textit{how aggressively} to adjust rankings for fairness, prioritizing adjustments where the model is least confident and deviation from CTR-optimal ordering is least costly.

\subsection{Design Inspiration from Algorithmic Trading}

\begin{figure}[H]
\centering
\vspace{-4pt}
\includegraphics[width=0.85\textwidth]{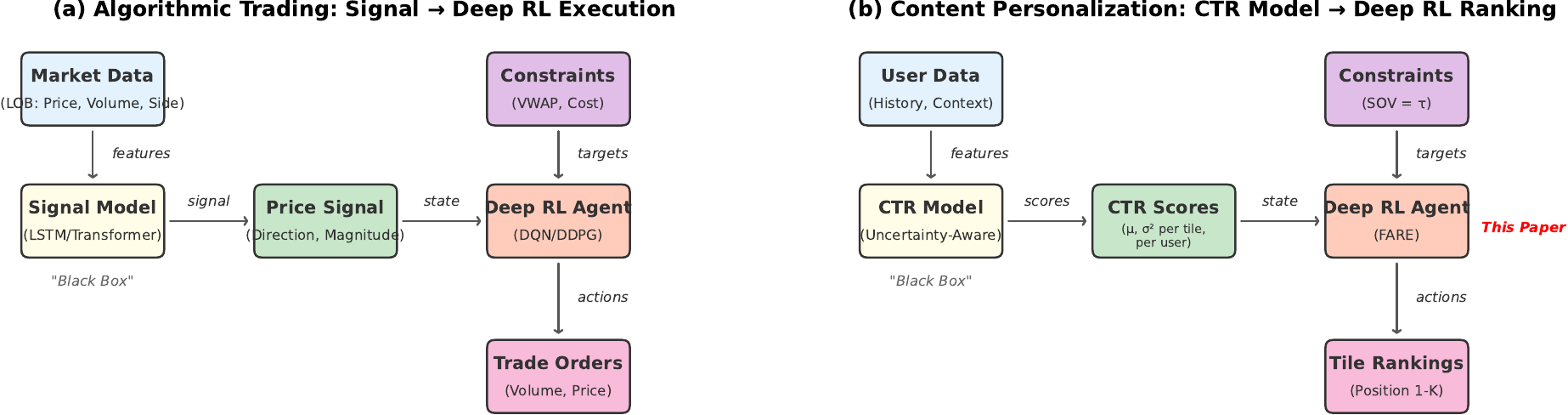}
\caption{Parallel architectures in (a) algorithmic trading and (b) content personalization, illustrating the signal$\rightarrow$execution decoupling pattern that FARE adapts for SOV-constrained re-ranking.}
\label{fig:pipeline}
\vspace{-8pt}
\end{figure}

A useful design parallel comes from algorithmic trading research, where a well-established pattern separates signal generation from execution optimization \citep{nagy2023asynchronous}: a signal model
produces predictions, while a separate execution agent, often a deep RL policy \citep{ning2021double}, translates those predictions into orders satisfying operational constraints such as Volume-Weighted Average Price (VWAP)\footnote{VWAP execution is a standard algorithmic trading objective: execute a large order over a time window such that the average fill price tracks the market's volume-weighted average, minimizing market impact.} targets.
We adapt this pattern to personalized re-ranking: the CTR model plays the role of the signal model, and FARE acts as the execution layer.
Figure~\ref{fig:pipeline} illustrates this parallel.

This design yields a practical insight: \textbf{FARE is an uncertainty-aware plug-in layer} that can be added to any existing CTR model without retraining.
Modifying such models to incorporate fairness constraints---e.g., by adding SOV terms to the training loss---requires retraining and revalidating a production model. FARE instead operates \textit{downstream} of the CTR model, in the spirit of post-processing approaches to fairness \citep{hardt2016equality}, though the fairness gain is not free: enforcing supplier fairness can trade against user satisfaction \citep{mehrotra2018towards}. It takes predictions $(\mu, \sigma^2)$ as input and outputs adjusted rankings that satisfy SOV constraints. This separation of concerns means: (1) CTR models can be improved or swapped without touching the fairness layer; (2) SOV targets can be adjusted dynamically; and (3) the same FARE framework works with any uncertainty-aware CTR model such as Thompson Sampling \citep{chapelle2011empirical}, deep ensembles \citep{lakshminarayanan2017simple}, or Bayesian methods \citep{blundell2015weight}.

\subsection{Contributions}
\label{sec:contributions}

Our contributions are twofold. \textbf{First}, we frame SOV-constrained ranking as a deep reinforcement learning problem, adapting the signal$\rightarrow$execution decoupling pattern from algorithmic trading \citep{nagy2023asynchronous}, and instantiate it as FARE with three execution policies: a proportional controller (FARE-PC), an evolution-strategy policy (FARE-ES), and a policy-gradient variant (FARE-PPO). \textbf{Second}, we use prediction uncertainty $\sigma^2$ as the \emph{gain} of that controller: the correction applied to a group is scaled by how uncertain the model is about it, so adjustments concentrate where deviation from CTR-optimal ordering is cheapest. Uncertainty has been used for exploration \citep{chapelle2011empirical, chinta2025variance} and to bound per-query re-ranking \citep{heuss2023predictive}, and exposure deficits have been controlled without it \citep{morik2020controlling, biega2018equity, xu2023pmmf}; using uncertainty as the gain of a cumulative-allocation controller is distinct from all three. We further find that neither learned policy is reliable across regimes---gradient-free search leads on synthetic data, policy gradients on KuaiRand-Pure---while the reactive controller outperforms both on single-objective fairness, which we attribute to weak per-step credit assignment.

\section{Problem Formulation}

\subsection{Setting}
\label{sec:setting}

A CTR model $f_\theta$ (treated as a black box) produces engagement predictions for each content item (referred to as a \textit{tile})\footnote{A tile refers to an individual content item (e.g., article, video, or advertisement) displayed at a specific ranked position in a personalised list.} $i \in \{1,\ldots,K\}$ and user $u$. Following standard uncertainty-aware approaches \citep{lakshminarayanan2017simple, chinta2025variance}, we assume the model outputs both a mean prediction $\mu_i(u)$ and variance $\sigma_i^2(u)$ capturing predictive uncertainty:
\begin{equation}
  \text{Score}_i(u) \sim \mathcal{N}(\mu_i(u), \sigma_i^2(u))
\end{equation}

Rankings are determined by sampling scores from these distributions and ordering by descending sampled values. This is standard practice in Thompson Sampling-based systems \citep{chapelle2011empirical}---and critically, our \textbf{CTR-Only baseline} is precisely this: sampling $s_i \sim \mathcal{N}(\mu_i, \sigma_i^2)$ for all tiles and ranking by descending $s_i$. CTR-Only therefore serves as the uncertainty-aware Thompson Sampling baseline that the literature recommends; FARE operates \textit{downstream} of this step by adjusting the means $\mu_i$ before sampling, without modifying the CTR model itself. A purely deterministic baseline would rank by descending $\mu_i$, maximising expected CTR but ignoring both uncertainty and fairness.

FARE is agnostic to the \emph{unit of allocation}: it enforces SOV over whatever unit the upstream CTR model scores. Where that model scores individual items, SOV is enforced over items; where it scores groups---content shelves, product categories, editorial modules---SOV is enforced over groups. The controller itself sees only a vector of $(\mu_i, \sigma_i^2)$ pairs and a per-unit exposure deficit, and is indifferent to what the index $i$ ranges over.  What our evaluation does fix is that the candidate set equals the slate---every candidate is shown, so the only decision is ordering, with $K \leq 10$ exposure groups throughout our experiments. FARE as evaluated does not address the setting where a large catalogue is filtered to a short slate, so that most items receive no exposure at all (Appendix~\ref{sec:discussion}).

\subsection{Share of Voice Constraint}
\label{sec:sov}

We define \textbf{Share of Voice (SOV)} for tile $i$ in position 1 over $N$ users and $T$ days:
\begin{equation}
  \text{SOV}_i = \frac{1}{NT} \sum_{d=1}^{T} \sum_{u=1}^{N} \mathbf{1}[\text{rank}_1(u, d) = i]
  \label{eq:sov}
\end{equation}
where $N$ is the number of users, $T$ is the number of days, and $\mathbf{1}[\text{rank}_1(u, d) = i]$ is an indicator function that equals 1 if tile $i$ appears in position 1 for user $u$ on day $d$, and 0 otherwise.

In general, the constraint is $\text{SOV}_i = \tau_i$ where $\tau_i$ can differ per tile. In the default case, we consider equal allocation where $\tau_i = \tau = 1/K$ for all tiles (e.g., 20\% for $K=5$), though FARE's deficit formula ($\text{deficit}_i = \tau_i - \text{SOV}_i^{(t)}$) handles arbitrary $\tau_i$ without modification, as validated in Section~\ref{sec:nonuniform}. In practice, targets may be set by contractual agreements with content providers or by business objectives such as proportional allocation to revenue contribution. Such targets are \emph{specified externally}, rather than derived from the model's own relevance estimates as in merit-proportional schemes \citep{morik2020controlling}, which we do not evaluate. Dynamic targets $\tau_i(t)$ (e.g., seasonal promotions) are a natural extension left to future work.

\subsection{Constrained Optimization Problem}

Analogous to constrained trade execution \citep{ning2021double}, we formulate:
\begin{equation}
\max_{\pi} \; \mathbb{E}\left[\sum_{d,u} \text{CTR}(\pi(u, d))\right] \quad \text{s.t.} \quad |\text{SOV}_i - \tau_i| \leq \epsilon, \; \forall i
\label{eq:constrained}
\end{equation}
where $\pi$ is the ranking policy, $\text{CTR}(\pi(u, d))$ is the click-through rate of the ranking produced by policy $\pi$ for user $u$ on day $d$, $\tau_i$ is the target SOV for tile $i$, and $\epsilon$ is the tolerance for SOV deviation. The policy $\pi$ must translate CTR signals into rankings that satisfy SOV constraints while minimizing CTR loss---exactly as a trade executor translates price signals into orders satisfying VWAP constraints while minimizing execution cost.

\section{Methodology}
\label{sec:methods}

\subsection{Input Generation}
\label{sec:input_gen}

FARE takes as input a mean prediction $\mu$ and a variance $\sigma^2$ per item, where $\mu$ represents the expected engagement score and $\sigma^2$ quantifies the uncertainty associated with that prediction. In practice, these can be sourced from an upstream CTR model or generated synthetically for evaluation purposes. To obtain these inputs to FARE, we employ two complementary approaches depending on the evaluation setting. In the first approach, we directly synthesise $(\mu, \sigma^2)$ using a hierarchical generative model with controllable properties, enabling systematic and reproducible evaluation across a range of experimental conditions (Section~\ref{sec:synthetic_gen}). In the second approach, we source interaction logs from the KuaiRand-Pure public dataset and train a multi-head Gaussian MLP using Gaussian negative log-likelihood to produce learned $(\mu, \sigma^2)$ predictions from user features---providing a realistic, data-driven source of uncertainty-aware CTR estimates (Section~\ref{sec:kuairand}).

\subsubsection{Approach 1: Synthetic CTR Estimate Generation}
\label{sec:synthetic_gen}

For systematic evaluation, we generate synthetic CTR estimates $(\mu, \sigma^2)$ that produce a controlled tile ordering---specifically, Tile~1 $>$ Tile~2 $>$ $\cdots$ $>$ Tile~$K$ in Position~1 CTR---using a hierarchical model that decomposes CTR into base tile-level values plus day, user, and interaction variations:
\begin{equation}
\mu_{d,u,i} = \bar{\mu}_i + \delta_d^{(i)} + \epsilon_u^{(i)} + \eta_{d,u}^{(i)}, \quad \sigma^2_{d,u,i} = \bar{\sigma}^2_i + \nu_u^{(i)}
\label{eq:synth}
\end{equation}
The base means $\bar{\mu}_i$ are set in decreasing order to enforce the tile ordering, while the base variances $\bar{\sigma}^2_i$ vary independently of them, so that uncertainty is not a proxy for rank; the three noise terms introduce day, user, and user--day variation. Appendix~\ref{app:synth_constants} lists all constants.

Figure~\ref{fig:param_variation} confirms the generative model produces the desired tile ordering: Tile~1 dominates Position~1 at ${\sim}28\%$ while Tile~5 receives only ${\sim}11.5\%$, establishing the fairness gap that FARE must close. User-level distributions show realistic heterogeneity around base values. The full generation procedure is given in Algorithm~\ref{alg:param_gen}, Appendix~\ref{app:algorithms}. These synthesised $(\mu_i, \sigma_i^2)$ pairs are the direct input to the FARE execution layer in the synthetic experiments, and the generator additionally serves as the training environment for FARE-ES and FARE-PPO (Section~\ref{app:mdp_formulation}).

\subsubsection{Approach 2: Gaussian MLP for Uncertainty-Aware CTR Estimation}
\label{sec:kuairand}

To evaluate FARE beyond synthetic data, we train a multi-head Gaussian MLP on real interaction logs from the \textbf{KuaiRand-Pure} dataset~\citep{gao2022kuairand}, an unbiased recommendation dataset with randomly exposed videos containing 27,285 users, 7,583 short videos, and 17~days of interactions.

\textbf{Data Preprocessing.} We aggregate interactions at the \textit{category level} to construct tiles. Each video's primary category is extracted from its tag metadata, and we select the top~5 Level-1 categories by total interaction count, mapping them to tiles $\{1, \ldots, 5\}$. For each (user, day, category) triple, we compute the daily CTR as the ratio of clicks to impressions, yielding a continuous target $\in [0, 1]$. Missing observations (user--category pairs with no impressions on a given day) are marked with a sentinel value and masked during training. User features (76~dimensions including demographics and activity levels) are standardized, with categorical columns (e.g., \texttt{user\_active\_degree}) label-encoded to integers. Data is split chronologically to prevent temporal leakage, with the final 4~days reserved for testing.
Appendix~\ref{app:tile_construction} gives the full tile-construction rule and the target definition.

\textbf{Multi-Head Gaussian MLP.} The model takes user features as input and outputs $(\mu_i, \sigma^2_i)$ for each of the $K{=}5$ tiles simultaneously via $K$~independent output heads:
\begin{equation}
\small
\text{Input}(76) \!\xrightarrow{\text{FC+BN+ReLU}}\! 128 \!\rightarrow\! 64 \!\rightarrow\! 32 \!\rightarrow\! \underbrace{(\mu_i, \sigma^2_i)}_{i=1 \ldots 5}
\label{eq:mlp}
\end{equation}
Each head produces a linear mean $\mu_i$ and a softplus-activated variance $\sigma^2_i = \log(1 + \exp(z_i))$, trained with masked Gaussian negative log-likelihood:
\begin{equation}
\mathcal{L} = \frac{1}{|\mathcal{M}|} \sum_{(d,u,i) \in \mathcal{M}} \frac{(y_{d,u,i} - \mu_i)^2}{2\sigma^2_i} + \frac{1}{2}\log \sigma^2_i
\label{eq:gnll}
\end{equation}
where $\mathcal{M}$ is the set of observed (non-missing) entries. The model has ${\sim}21$K parameters, regularized with dropout (0.2) and early stopping (patience~150). The trained model's test-set predictions serve as direct input to the FARE execution layer. For the learning-based variants (FARE-ES, FARE-PPO), the trained MLP additionally acts as the interaction simulator during training: at each step, the policy receives the MLP's $(\mu_i, \sigma_i^2)$ predictions for the current (user, day) pair as its state, takes an action (score adjustments), and receives a reward computed from the resulting SOV deviation and position-weighted CTR loss (PWCL, defined in Section~\ref{sec:metrics}). This replay-based simulation does not require online user interactions---the learned MLP serves as a differentiable surrogate environment that models the distribution of real CTR patterns from the training logs.

\textbf{Prediction Characteristics.} Figure~\ref{fig:kuairand_params} shows the learned $(\mu, \sigma^2)$ distributions. Unlike synthetic data, the model predictions exhibit: (1)~\textit{non-monotonic tile ordering}---Tiles~1 and~3 both dominate Position~1 with similar SOV (${\sim}23\%$), while Tiles~2 and~4 are near-equal at ${\sim}19\%$; (2)~\textit{narrower baseline SOV spread}---the gap between highest and lowest SOV (23.3\% vs.\ 15.1\%) is roughly half that of the synthetic setup (28.0\% vs.\ 11.5\%); and (3)~\textit{small predicted variances}---median $\sigma^2 \approx 0.03$, an order of magnitude smaller than $\mu$, reflecting low predicted variability in daily CTR.

\subsection{FARE Architecture}

FARE is an uncertainty-aware plug-in layer atop any CTR model (see Figure~\ref{fig:pipeline}). The CTR model (e.g., with uncertainty quantification) is frozen; FARE adjusts its outputs to satisfy SOV constraints. We implement three execution policies: \textbf{FARE-PC} (control-theoretic baseline), \textbf{FARE-ES} (evolution strategy, derivative-free policy search), and \textbf{FARE-PPO} (Proximal Policy Optimization, gradient-based deep reinforcement learning (RL)).

The SOV target $\tau$ does not enter the CTR model's training objective at any point. The multi-head Gaussian MLP of Section~\ref{sec:kuairand} is trained purely with masked Gaussian negative log-likelihood (Eq.~\ref{eq:gnll}), is frozen before FARE runs, and receives no gradient from the SOV term; SOV enters only downstream, as the deficit in FARE-PC's control law and as one term of the reward for FARE-ES and FARE-PPO.

\textbf{Explicit vs.\ Implicit Use of Uncertainty.} The three FARE variants use uncertainty differently: FARE-PC \textit{explicitly} multiplies by $\sigma$ in its adjustment formula ($\text{adj} \propto \sigma / \text{remaining}$), guaranteeing that high-uncertainty tiles receive larger, lower-cost adjustments. FARE-ES and FARE-PPO receive $\sigma^2$ as \textit{input features} and may learn similar or different strategies---more flexible but less interpretable.

\textbf{Common Pipeline.} All three FARE variants share the same end-to-end flow, taking $(\mu_i, \sigma_i^2)$ as input---obtained via either approach described in Section~\ref{sec:input_gen}:
\begin{equation}
\small
\boxed{
\mathcal{N}(\mu_i, \sigma_i^2)
\!\xrightarrow{\text{\tiny FARE}}\!
\mathcal{N}(\tilde{\mu}_i, \sigma_i^2)
\!\xrightarrow{\text{\tiny sample}}\!
\tilde{s}_i
\!\xrightarrow{\text{\tiny rank}}\!
\text{sort}_{i}^{\downarrow}(\tilde{s}_i)
}
\label{eq:pipeline}
\end{equation}
The key insight is that FARE only adjusts the \textit{mean} ($\mu_i \rightarrow \tilde{\mu}_i$); the variance $\sigma_i^2$ remains unchanged. After adjusting the mean, scores are sampled from $\mathcal{N}(\tilde{\mu}_i, \sigma_i^2)$ and tiles are ranked by descending sampled score. This preserves the uncertainty structure of the original CTR model, keeping FARE consistent with the Thompson Sampling behaviour of the CTR-Only baseline.

The three variants differ only in \textit{how} the adjustment $\tilde{\mu}_i - \mu_i$ is computed:
\begin{itemize}
    \item \textbf{FARE-PC}: Explicit formula based on SOV deficit and uncertainty (no learning)
    \item \textbf{FARE-ES}: Neural network trained via evolution strategy
    \item \textbf{FARE-PPO}: Neural network trained via policy gradients
\end{itemize}

\label{app:mdp_formulation}
\textbf{Markov Decision Process (MDP) Formulation.} While FARE-PC uses feedback control without learning, the learning-based methods (FARE-ES and FARE-PPO) formulate the problem as a Markov Decision Process \citep{sutton2018reinforcement} with the following components:

\begin{description}[leftmargin=1.5em, labelindent=0pt]
\item[\textbf{State} $X_t$:] Concatenation of CTR predictions (with uncertainty), current SOV, SOV deficit, and time remaining:
\begin{equation}
\begin{split}
  X_t = [&\mu_1, \ldots, \mu_K,\; \sigma_1^2, \ldots, \sigma_K^2, \\
  &\text{SOV}_1^{(t)}, \ldots, \text{SOV}_K^{(t)},\; \Delta\text{SOV}_1^{(t)}, \ldots, \Delta\text{SOV}_K^{(t)},\\
  &(T{-}t)/T]
\end{split}
\end{equation}
where $\Delta\text{SOV}_i^{(t)} = \tau_i - \text{SOV}_i^{(t)}$ is the deficit from target (using $\tau_i = \tau$ for equal allocation). We include both SOV and $\Delta$SOV in the state to naturally support non-uniform targets $\tau_i$ across tiles (evaluated in Section~\ref{sec:nonuniform}).

\item[\textbf{Action} $a_t \in \mathbb{R}^K$:] Score adjustments applied \textit{additively}:
\begin{equation}
  \tilde{\mu}_i = \mu_i + \alpha \cdot a_t^{(i)}, \quad a_t^{(i)} \in [-1, 1]
\end{equation}
where $\alpha > 0$ controls the maximum adjustment magnitude. We use additive (not multiplicative) adjustments for two reasons: (1) they provide equal absolute boosts regardless of base CTR, enabling low-$\mu$ tiles to compete for Position~1 without requiring the adjustment to scale with the tile's own mean; and (2) they preserve the original rank ordering when all adjustments are zero, ensuring the CTR-only policy is recovered as a special case ($\alpha \to 0$). Multiplicative scaling would amplify boosts for already-high-$\mu$ tiles and suppress boosts for low-$\mu$ tiles, counteracting the fairness objective. Following the common pipeline (Eq.~\ref{eq:pipeline}), scores are then sampled from $\mathcal{N}(\tilde{\mu}_i, \sigma_i^2)$ and ranked to determine the winner.

\item[\textbf{Reward}:] Balancing SOV satisfaction and CTR preservation (analogous to execution cost vs VWAP deviation in constrained MDPs \citep{altman1999constrained}):
\begin{equation}
  r_t = -\lambda_{\text{sov}} \sum_{i=1}^{K} (\text{SOV}_i^{(t)} - \tau_i)^2 - \lambda_{\text{ctr}} \cdot \text{PWCL}_t
  \label{eq:reward}
\end{equation}
where $\text{PWCL}_t = \sum_j w_j \times (s_{[j]} - \tilde{s}_{[j]})$ is the Position-Weighted CTR Loss---a metric we introduce to quantify engagement degradation due to re-ranking, with $s_{[j]}$ and $\tilde{s}_{[j]}$ the sampled scores of the tiles at position $j$ under CTR-Only and under FARE, and DCG-style position weights $w_j$ \citep{jarvelin2002cumulated}. The first term penalizes deviation from SOV targets; the second term penalizes CTR degradation weighted by position importance. The weights $\lambda_{\text{sov}}$ and $\lambda_{\text{ctr}}$ control the fairness--engagement trade-off. These weights affect the learning-based variants (FARE-ES, FARE-PPO) but not FARE-PC, which uses no reward signal.
\end{description}

\textbf{Training Simulator.} For FARE-ES and FARE-PPO in the synthetic experiments, the MDP environment is constructed directly from the synthetic parameter generator; in the KuaiRand-Pure experiments the trained Gaussian MLP of Section~\ref{sec:kuairand} plays the same role. At each step, the environment samples $(\mu_i, \sigma_i^2)$ for the current (user, day) pair, the agent outputs score adjustments $a_t$, scores are sampled from $\mathcal{N}(\tilde{\mu}_i, \sigma_i^2)$ to determine the winner, and the reward is computed from the resulting SOV deviation and PWCL. Episodes span all $N \times T_{\text{train}}$ impressions, with SOV counts reset between episodes. This replay-based simulator does not require online user interactions---the generative model acts as a surrogate environment, making the training setup self-contained. FARE-PC requires no training environment as it is a reactive control policy.

\subsubsection{FARE-PC: Uncertainty-Weighted Proportional Control}
\label{sec:farepc}

Unlike the learning-based methods, FARE-PC explicitly incorporates the uncertainty $\sigma_i$ to scale adjustments---tiles with higher uncertainty receive larger adjustments since we have more ``freedom'' to deviate from predictions we are less confident about. FARE-PC is a \textit{proportional controller}: it applies an adjustment proportional to the current error (the SOV deficit). Here the ``plant'' is the cumulative SOV, and the ``disturbance'' is the stochastic variation in CTR sampling. The deficit term $\text{deficit}_i = \tau_i - \text{SOV}_i^{(t)}$ is both the error signal and the self-correction mechanism: a positive deficit boosts the tile until it meets its target, and a negative deficit applies a penalty until excess exposure is absorbed.

Algorithm~\ref{alg:vwap} in Appendix~\ref{app:algorithms} states the full procedure.

\textbf{State tracking} (no future lookahead): $\text{position\_counts}_i$ tracks tiles placed in position 1 so far (past), $t$ is the total impressions so far (past), $(\mu_i, \sigma_i)$ are the current user's CTR predictions with uncertainty (present), and $T_{\text{total}}$ is the known constraint horizon (given).

\textbf{Deficit calculation} (how far from target SOV):
\begin{equation}
  \text{deficit}_i = \tau - \text{SOV}_i^{(t)} = \tau - \frac{\text{position\_counts}_i}{t}
\end{equation}

Note that $\text{deficit}_i$ is computed \textit{per tile} and can be positive or negative:
\begin{itemize}
  \item \textbf{Positive deficit} ($\text{SOV}_i < \tau$): Tile is \textit{behind} target $\rightarrow$ receives a \textit{boost}
  \item \textbf{Negative deficit} ($\text{SOV}_i > \tau$): Tile is \textit{ahead} of target $\rightarrow$ receives a \textit{penalty}
\end{itemize}
This ensures tiles with excess exposure are penalized while underexposed tiles are boosted, naturally balancing SOV across all tiles.

\textbf{Time remaining factor} (increases aggressiveness as deadline approaches):
\begin{equation}
  \text{remaining} = T_{\text{total}} - t
\end{equation}
Early in the horizon, $1/\text{remaining}$ is small (gentle adjustments); near the deadline, it grows large (aggressive adjustments).

\textbf{Convergence.} The deficit formula is self-correcting by construction: a positive deficit boosts a tile's adjusted mean, increasing its probability of winning and reducing the deficit over subsequent impressions. In practice, convergence is governed by $\alpha$: with uniform targets and $K = 5$, $\alpha = 2.0$ brings SOV close to target over the 10-day test period (Section~\ref{sec:alpha_ablation}), whereas non-uniform targets need $\alpha \approx 3.0$ and residual error grows with $K$ (Section~\ref{sec:scalability_K}). A proportional controller on a linear integrating plant would reach zero steady-state error \citep{astrom2008feedback}, but the correspondence is not exact here---the map from adjusted means to win probabilities is nonlinear and the gain grows as the horizon closes---so we do not claim guaranteed convergence: residual error remains at larger $K$ and under DCG weighting, and the controller diverges when a second objective is weighted heavily (Section~\ref{sec:multiobj}). If CTR predictions drift over time (e.g., due to model updates), the feedback loop continues to self-correct but the convergence point may shift.

\textbf{Score adjustment} (uncertainty-weighted):
\begin{equation}
\label{eq:farepc}
  \tilde{\mu}_i = \mu_i + \alpha \cdot \text{deficit}_i \cdot \sigma_i / \text{remaining}
\end{equation}

The key insight is that $\sigma_i$ scales the adjustment in \textit{both directions}: tiles with high uncertainty that are behind target get large boosts, while tiles with high uncertainty that are ahead of target get large penalties. This is justified because we have less confidence in these predictions, giving us more freedom to adjust them for fairness.

Following the common pipeline (Eq.~\ref{eq:pipeline}), scores are sampled from $\mathcal{N}(\tilde{\mu}_i, \sigma_i^2)$ to determine the final ranking.

\subsubsection{FARE-ES: Evolution Strategy Policy}
\label{sec:farees}

Using the MDP formulation above, we train a neural network policy using evolution strategy (ES) \citep{salimans2017evolution}. Unlike policy gradient methods, ES updates weights through population-based search without backpropagation:

Algorithm~\ref{alg:fare_es} in Appendix~\ref{app:algorithms} gives the training loop.

\textbf{Policy}: 2-layer MLP with ReLU activation: $\pi_\theta: X \rightarrow a \in [-1,1]^K$

\textbf{Training}: Evolution strategy with population size 10 (mirrored sampling) over 100 episodes. At each generation, we perturb the policy weights, evaluate fitness (cumulative reward over entire trajectory), and update toward better-performing perturbations.

\subsubsection{FARE-PPO: Proximal Policy Optimization}
\label{sec:fareppo}

Using the same MDP formulation, we implement a Proximal Policy Optimization (PPO) agent \citep{schulman2017proximal} with a combined actor-critic network. PPO is a popular deep reinforcement learning algorithm that uses clipped surrogate objectives to ensure stable policy updates.

Algorithm~\ref{alg:fare} in Appendix~\ref{app:algorithms} gives the training loop.

\textbf{Policy}: Combined actor-critic network with a shared 2-layer MLP (128 hidden units), an actor head outputting action mean $a \in [-1,1]^K$ with learned log-std, and a critic head outputting value estimate $V(X)$.

\textbf{Training}: We use PPO with clipped surrogate objective ($\epsilon=0.2$), Generalized Advantage Estimation (GAE, $\lambda=0.95$) \citep{schulman2016high}, 4 PPO epochs per trajectory, entropy bonus for exploration, and Adam optimizer.

\subsection{Computational Complexity}
\label{sec:complexity}

FARE-PC has the lowest inference and memory cost ($O(K)$ per impression, no stored model weights) and requires only one tuned hyperparameter ($\alpha$), making it the natural choice for latency-sensitive production deployments. FARE-ES and FARE-PPO require one forward pass through a small MLP per impression, but differ substantially in training: FARE-ES evaluates $P$ complete trajectories per generation to produce a single parameter update, while FARE-PPO performs several gradient epochs on each trajectory.

\section{Experiments}

We evaluate FARE in two settings: (1)~synthetic data with controlled properties---uniform and non-uniform SOV targets, performance across increasing $K \in \{3, 5, 7, 10\}$ tiles, multi-position SOV weighting, and multi-objective SOV + diversity optimization, and (2)~evaluation on KuaiRand-Pure with learned model predictions, including uniform targets, non-uniform targets, and multi-position SOV weighting.

\subsection{Experimental Configuration}
\label{sec:exp_config}

The synthetic setting uses $N=10{,}000$ users, $T=40$ days and $K=5$ tiles (Section~\ref{sec:synthetic_gen}), with Days~1--30 for training the learned variants and Days~31--40 for testing (100,000 test impressions). The KuaiRand-Pure test set contains 4~days $\times$ 8,614~users $\times$ 5~tiles (172,280 impressions). All experiments use 3 random seeds; we report mean $\pm$ std. Key hyperparameters are as follows. FARE-PC uses $\alpha=2.0$; FARE-ES uses $\alpha=0.3$, $\lambda_{\text{sov}}=1.0$, $\lambda_{\text{ctr}}=0.1$, population size 10, 100 episodes; FARE-PPO uses $\alpha=0.5$, $\lambda_{\text{sov}}=2.0$, $\lambda_{\text{ctr}}=0.05$, 200 episodes.

\subsection{Evaluation Metrics}
\label{sec:metrics}

We evaluate methods along three dimensions: (1)~\textbf{Fairness}, measured by SOV Error ($\sum_i |\text{SOV}_i - \tau_i|$); (2)~\textbf{Engagement}, measured by PWCL using DCG-style position weights \citep{jarvelin2002cumulated}; and (3)~\textbf{Ranking Stability}, measured by Position Displacement (average positions moved per tile), Kendall Tau Distance (fraction of tile pairs in different relative order), and Top-1 Change Rate (fraction of impressions where the position-1 winner changes).

Because ground-truth clicks are not available for counterfactual rankings---no user was shown the re-ranked slate---PWCL is a model-based estimate of utility loss rather than a measured change in engagement. Appendix~\ref{app:metric_relations} relates SOV Error and PWCL to the exposure and accuracy measures used elsewhere in the literature.

\subsection{Performance Comparison}
\label{sec:perf_comparison}
\label{sec:baselines}

We compare FARE against three reference policies.
\textbf{CTR-Only} (Thompson Sampling) is the no-fairness reference.
\textbf{Max-Deficit} always promotes the most-underexposed tile to position~1, ignoring CTR entirely---a fairness upper bound and utility lower bound.
\textbf{Quota-CTR} follows the CTR-Only ranking but demotes over-quota tiles to the end of the slate, a hard-quota enforcement in the spirit of exposure-constrained re-ranking \citep{singh2018fairness, biega2018equity}.
These three span the extremes of the trade-off (no enforcement, hard enforcement, enforcement at any cost) and isolate the contribution of uncertainty weighting, since Quota-CTR and FARE-PC differ precisely in whether the fairness intervention is graduated and $\sigma$-scaled. These are reference points rather than published fair re-ranking systems; our claims are scoped accordingly, and Appendix~\ref{sec:limitations} sets out which methods are absent and what a like-for-like comparison would require.

\subsubsection{Uniform SOV Targets}
\label{sec:uniform_results}
\label{sec:synth_results}

We first evaluate under uniform targets ($\tau_i = 1/K = 20\%$ for all tiles), as defined in Section~\ref{sec:sov}.

\paragraph{Synthetic.}
Table~\ref{tab:baselines} reports results on the synthetic test set (mean $\pm$ std over 3 seeds). Max-Deficit achieves perfect SOV equality but at the highest engagement cost (11.74\% PWCL), since it ignores CTR entirely. Quota-CTR achieves 70.4\% reduction at 9.89\% PWCL by excluding over-quota tiles from contention. FARE-PC achieves 90.3\% SOV reduction at 2.64\% PWCL, reaching roughly one-third the SOV error of Quota-CTR at a fraction of the engagement cost (Table~\ref{tab:baselines}).

\begin{table}[H]
\centering
\caption{Synthetic test set results (mean $\pm$ std over 3 seeds). Target SOV: 20\% per tile. Upper block: non-FARE baselines. Lower block: FARE variants.
FARE-PC's variation across seeds is negligible on PWCL and the ranking-stability columns and is omitted; its SOV Error std rounds to 0.000.}
\label{tab:baselines}
\label{tab:results}
\resizebox{\textwidth}{!}{%
\begin{tabular}{l|c|c|c|c|c|c}
\toprule
\textbf{Method} & \textbf{SOV Error} $\downarrow$ & \textbf{Reduction} $\uparrow$ & \textbf{PWCL\%} $\downarrow$ & \textbf{Pos.\ Disp.} & \textbf{Kendall $\tau$} & \textbf{Top-1 Chg} \\
\midrule
CTR-Only (= TS) & 0.247 $\pm$ 0.002 & --   & 0.00\%       & 0.00 & 0.00 & 0.0\% \\
Max-Deficit   & 0.000 $\pm$ 0.000 & 100.0\% & 11.74 $\pm$ 0.29\% & --  & --  & --  \\
Quota-CTR    & 0.073 $\pm$ 0.008 & 70.4\% & 9.89 $\pm$ 0.11\%  & 1.39 $\pm$ 0.01 & 0.42 $\pm$ 0.00 & 23.4 $\pm$ 1.3\% \\
\midrule
\textbf{FARE-PC} & \textbf{0.024 $\pm$ 0.000} & \textbf{90.3\%} & \textbf{2.64\%} & 1.58 & 0.49 & 79.3\% \\
FARE-ES     & 0.034 $\pm$ 0.004 & 86.2\% & 2.79 $\pm$ 0.2\%  & 1.58 $\pm$ 0.001 & 0.49 $\pm$ 0.001 & 79.4 $\pm$ 0.1\% \\
FARE-PPO     & 0.155 $\pm$ 0.080 & 37.2\% & 1.50 $\pm$ 0.1\%  & 1.56 $\pm$ 0.02 & 0.48 $\pm$ 0.01 & 78.5 $\pm$ 0.7\% \\
\bottomrule
\end{tabular}%
}
\end{table}

The uncertainty-weighted design explains FARE-PC's efficiency advantage: rather than always forcing the most-underexposed tile (Max-Deficit), it applies graduated score adjustments proportional to the SOV deficit and uncertainty, allowing the CTR-ordering signal to persist for high-confidence predictions while prioritising fairness corrections for uncertain ones. FARE-ES near-matches the hand-designed controller; FARE-PPO improves only moderately and with high variance. We attribute this to weak per-step credit assignment; Appendix~\ref{app:ppo_failure} discusses the mechanism. Table~\ref{tab:sov_distribution} shows that FARE-PC and FARE-ES bring all tiles within 1 percentage point of the 20\% target, while FARE-PPO leaves Tile~5 at 15.9\%.

\begin{table}[H]
\centering
\begin{minipage}[c]{0.56\textwidth}
\centering
\caption{Final SOV distribution on test set (\%), one representative seed; T1--T5 denote the five tiles.}
\label{tab:sov_distribution}
\resizebox{\textwidth}{!}{%
\scriptsize
\begin{tabular}{l|ccccc|c}
\toprule
 & \textbf{T1} & \textbf{T2} & \textbf{T3} & \textbf{T4} & \textbf{T5} & \textbf{Error} \\
\midrule
CTR-Only & 28.0 & 23.6 & 20.7 & 16.1 & 11.5 & 0.25 \\
FARE-PC & 20.9 & 20.2 & 19.9 & 19.6 & 19.3 & 0.02 \\
FARE-ES & 19.9 & 20.6 & 20.1 & 19.6 & 19.9 & 0.03 \\
FARE-PPO & 24.9 & 19.7 & 21.3 & 18.1 & 15.9 & 0.16 \\
\midrule
Target & 20.0 & 20.0 & 20.0 & 20.0 & 20.0 & 0.00 \\
\bottomrule
\end{tabular}%
}
\end{minipage}
\hfill
\begin{minipage}[c]{0.40\textwidth}
\centering
\includegraphics[width=0.75\linewidth]{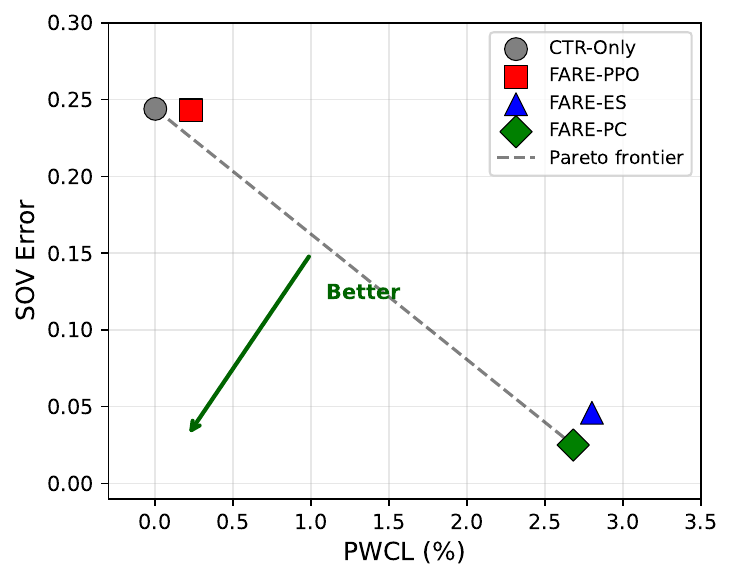}
\captionof{figure}{Pareto frontier.}
\label{fig:pareto}
\end{minipage}
\end{table}

The Pareto frontier (Figure~\ref{fig:pareto}) confirms FARE-PC dominates: it achieves the best SOV correction at the lowest engagement cost.
FARE-ES settles at a higher final SOV Error (0.034) than FARE-PC (0.024), and FARE-PPO ends with a ${\sim}9$ percentage point spread across tiles (Table~\ref{tab:sov_distribution}), indicating incomplete optimization.

\paragraph{Gaussian NLL MLP (KuaiRand-Pure).}
Non-monotonic tile orderings and milder baseline imbalance (SOV Error 0.122 vs.\ 0.247 synthetic) make this a different regime from the synthetic setting.

\begin{table}[H]
\centering
\caption{KuaiRand-Pure evaluation metrics (mean $\pm$ std). Target SOV: 20\% per tile.
}
\label{tab:kuairand_metrics}
\resizebox{\textwidth}{!}{%
\begin{tabular}{l|c|c|c|c|c|c}
\toprule
\textbf{Method} & \textbf{SOV Error} $\downarrow$ & \textbf{Reduction} $\uparrow$ & \textbf{PWCL\%} $\downarrow$ & \textbf{Pos.\ Disp.} & \textbf{Kendall $\tau$} & \textbf{Top-1 Chg} \\
\midrule
CTR-Only & 0.122 $\pm$ 0.005 & -- & 0.0\% & 0.00 & 0.00 & 0.0\% \\
FARE-PC & \textbf{0.010 $\pm$ 0.001} & \textbf{91.8\%} & 0.19 $\pm$ 0.01\% & 1.59 $\pm$ 0.003 & 0.497 $\pm$ 0.001 & 79.8 $\pm$ 0.2\% \\
FARE-ES & 0.099 $\pm$ 0.009 & 18.9\% & 0.06 $\pm$ 0.01\% & 1.59 $\pm$ 0.004 & 0.494 $\pm$ 0.001 & 79.5 $\pm$ 0.2\% \\
FARE-PPO & 0.089 $\pm$ 0.032 & 27.0\% & 0.10 $\pm$ 0.08\% & 1.59 $\pm$ 0.006 & 0.495 $\pm$ 0.003 & 79.6 $\pm$ 0.3\% \\
\bottomrule
\end{tabular}%
}
\end{table}

FARE-PC achieves 91.8\% SOV error reduction (0.122 $\rightarrow$ 0.010), closely matching its synthetic performance (90.3\%) and confirming that proportional control generalizes to noisy parameter estimates (Table~\ref{tab:kuairand_metrics}). Table~\ref{tab:kuairand_sov} shows the per-tile SOV distribution: FARE-PC brings all tiles within 0.4~percentage points of the 20\% target. Notably, the relative ordering shifts: FARE-PPO (27.0\% reduction) outperforms FARE-ES (18.9\%) on KuaiRand-Pure data, reversing the synthetic ranking---the milder baseline imbalance benefits PPO's moderate adjustments while ES overshoots. FARE-PC, being reactive rather than learned, adapts naturally to either regime.

\subsubsection{Non-Uniform SOV Targets}
\label{sec:nonuniform}

\paragraph{Synthetic.}
We evaluate FARE under asymmetric per-tile targets $\tau = [10, 15, 20, 25, 30]\%$---a substantially harder problem (baseline SOV Error 0.547 vs.\ 0.247 uniform) requiring the system to \textit{reverse} the natural CTR ordering. Tables~\ref{tab:nonuniform_metrics} and~\ref{tab:nonuniform_sov} in Appendix~\ref{app:nonuniform_results} report the results. FARE-PC and FARE-ES both achieve ${\sim}78\%$ error reduction without algorithmic modification, while FARE-PPO achieves only 32.9\% with high variance ($\pm$0.144). The near-parity of FARE-PC and FARE-ES under non-uniform targets (gap of 0.7pp vs.\ 4.1pp under uniform) suggests that larger SOV deficits provide a stronger training signal for FARE-ES.

\paragraph{Gaussian NLL MLP (KuaiRand-Pure).}
Under non-uniform targets ($\tau = [10, 15, 20, 25, 30]\%$), FARE-PC and FARE-ES achieve comparable reductions (83.0\% and 81.6\%), mirroring the synthetic finding.

\subsubsection{Multi-Position SOV Weighting}
\label{sec:multipos_sov}

We generalize SOV from Top-1 to Top-3 Uniform and DCG-Weighted schemes (Eq.~\ref{eq:multipos_sov}), where $w_p$ are position weights. The weighting scheme fundamentally changes problem difficulty: baseline error ranges from 0.247 (Top-1) to 0.061 (DCG).

\begin{equation}
\text{SOV}_i = \frac{\sum_{u,t} \sum_{p=1}^{K} w_p \cdot \mathbf{1}[\text{tile}_i \text{ at position} p]}{\sum_{u,t} \sum_{p=1}^{K} w_p}
\label{eq:multipos_sov}
\end{equation}

\paragraph{Synthetic.}

FARE-PC is the only method that improves fairness under all three schemes (Table~\ref{tab:multipos_sov}, Appendix~\ref{app:multipos_results}). Under DCG weighting, both FARE-ES ($-$9.8\%) and FARE-PPO ($-$3.3\%) \textit{degrade} fairness---the weak baseline signal (0.061) prevents effective learning. Notably, FARE-ES outperforms FARE-PC under Top-3 (73.7\% vs.\ 70.8\%), benefiting from the smoother optimization landscape.

\paragraph{Gaussian NLL MLP (KuaiRand-Pure).}

FARE-PC maintains consistent performance across all weighting schemes (Table~\ref{tab:kuairand_multipos}) (91.8\%, 72.6\%, 49.6\% reduction)---closely matching the synthetic results. Unlike the synthetic setting, learned methods no longer fail under DCG weighting: FARE-PPO achieves a 16.7\% reduction, and it also outperforms FARE-ES under Top-3 (25.4\% vs.\ 12.7\%).

\subsubsection{Multi-Objective: SOV + Category Diversity}
\label{sec:multiobj}

We extend FARE to jointly optimize SOV fairness and category diversity by adding a diversity term: $\tilde{\mu}_i = \mu_i + \Delta_i^{\text{SOV}} + \lambda_{\text{div}} \cdot \Delta_i^{\text{div}}$, where $\Delta_i^{\text{SOV}}$ is the variant's SOV adjustment (Eq.~\ref{eq:farepc} for FARE-PC, $\alpha \cdot a_t^{(i)}$ for the learned variants) and $\Delta_i^{\text{div}}$ is the diversity adjustment; scores are then sampled from $\mathcal{N}(\tilde{\mu}_i, \sigma_i^2)$ as in Eq.~\ref{eq:pipeline}. Tiles are assigned content categories (News, Finance, Sports) and diversity measures distinct categories in the top-$k$ positions. We sweep $\lambda_{\text{div}} \in \{0, 0.5, 1.0, 2.0, 5.0\}$.

\paragraph{Synthetic.}
This experiment exposes the Achilles' heel of hand-designed control (Table~\ref{tab:multiobj}, Appendix~\ref{app:multiobj_results}). FARE-PC's SOV Error increases $26\times$ (0.024 $\rightarrow$ 0.632) as $\lambda_{\text{div}}$ rises from 0 to 5---collapsing worse than the CTR-Only baseline---because the rigid additive heuristic cannot balance competing score adjustments. FARE-PC reaches its best SOV (0.010) at $\lambda_{\text{div}}{=}0.5$ before deteriorating steeply. In contrast, FARE-ES degrades only $2.6\times$ (0.034 $\rightarrow$ 0.088), with its learned policy naturally discovering the joint optimization landscape in a smooth, monotonic fashion. At $\lambda_{\text{div}}{=}2.0$, FARE-ES dominates FARE-PC on the primary fairness metric (0.063 vs.\ 0.175). This identifies a clear deployment guideline: use FARE-PC when the fairness objective dominates, and FARE-ES once a competing objective carries substantial weight ($\lambda_{\text{div}} \geq 2$ here), where FARE-PC's SOV Error degrades sharply while FARE-ES remains stable.

\paragraph{Gaussian NLL MLP (KuaiRand-Pure).}The diversity term is degenerate in this setting, so we do not repeat the sweep: the five tiles correspond one-to-one to the five Level-1 categories, so any top-$k$ slate contains exactly $k$ distinct categories and the objective is constant. The synthetic setting admits a non-trivial trade-off because its five tiles are mapped onto three content categories; recovering one here would require a coarser grouping over the categories than the preprocessing produces.

\subsection{Ablation Study}
\label{sec:ablation}

\subsubsection{Effect of Increasing the Number of Tiles}
\label{sec:scalability_K}

We sweep $K \in \{3, 5, 7, 10\}$ with uniform targets. Table~\ref{tab:scalability_K} reveals a clear robustness hierarchy: FARE-PC degrades gracefully (94.4\% $\rightarrow$ 77.1\% reduction) with stable PWCL (2.45\% $\rightarrow$ 2.87\%), FARE-ES follows comparably (88.1\% $\rightarrow$ 73.9\%) but with increasing variance, and FARE-PPO collapses at $K{=}10$ (0.8\% reduction, i.e.\ no material improvement over the no-fairness baseline)---consistent with policy-gradient variance growing with the number of action components, as it does with episode length \citep{salimans2017evolution}, here from $\mathbb{R}^3$ to $\mathbb{R}^{10}$. FARE-PC's control law degrades gracefully because each tile's adjustment is computed independently based on its deficit.

\subsubsection{Sensitivity Analysis: Control Gain $\alpha$}
\label{sec:alpha_ablation}

The gain parameter $\alpha$ governs the aggressiveness of FARE-PC's score adjustments. We sweep $\alpha \in \{0.5, 1.0, 2.0, 3.0, 5.0\}$ under both uniform and non-uniform targets. Figure~\ref{fig:alpha_uniform} shows the uniform case: SOV Error drops sharply from 0.125 at $\alpha{=}0.5$ to 0.024 at $\alpha{=}2.0$, while PWCL rises from 1.4\% to 2.6\% with diminishing marginal cost. The operating point $\alpha{=}2.0$ lies at the Pareto curve knee. Under non-uniform targets, the same pattern holds but with higher absolute costs ($\alpha \approx 3.0$ for near-convergence). In practice, $\alpha$ can be tuned to the operator's tolerance for CTR degradation---a deployment advantage of the interpretable control approach.

\subsubsection{Robustness to Uncertainty Miscalibration}
\label{sec:robustness_sigma}

A key practical concern is whether FARE degrades when the upstream model's uncertainty estimates $\sigma^2$ are miscalibrated---either systematically too small (over-confident model) or too large (under-confident model). We evaluate this by multiplying all $\sigma^2$ values by a scale factor $s \in \{0.25\times, 0.5\times, 1\times, 2\times, 4\times\}$, covering a 16$\times$ range. CTR-Only (Thompson Sampling) serves as the no-fairness reference at each scale. FARE-PC consistently reduces SOV error relative to CTR-Only across the entire range, confirming robustness regardless of whether the upstream model is over- or under-confident. The gap over CTR-Only does not collapse across a 16$\times$ range of $\sigma^2$ magnitudes (Figure~\ref{fig:sigma_sensitivity}). The absolute SOV error also falls as $s$ grows, but this is a gain effect rather than a benefit of miscalibration: $\sigma_i$ multiplies the correction in Eq.~\ref{eq:farepc}, so inflating it is equivalent to raising $\alpha$, and should carry an analogous PWCL cost to the one documented in Section~\ref{sec:alpha_ablation}. The robustness claim is the persistence of the gap, not the level of the curve.

\section{Related Work}
\label{sec:related}

\textbf{Deep RL for Trade Execution.} Prior work uses deep RL for trade execution \citep{ning2021double}, and \citet{nagy2023asynchronous} separate signal forecasting from order execution---a paradigm we adapt for SOV-constrained content re-ranking.

\textbf{Uncertainty in Reinforcement Learning.} Uncertainty in RL usually concerns the RL system itself: value functions for exploration \citep{osband2016deep}, return distributions \citep{bellemare2017distributional}, or dynamics models for planning \citep{chua2018deep}. FARE instead uses an \textit{external upstream predictor's} $\sigma^2$ as a state feature for execution.

\textbf{Uncertainty-Aware Ranking.} Thompson Sampling \citep{chapelle2011empirical} and Upper Confidence Bound (UCB) methods \citep{auer2002finite} are widely used for exploration in recommendations. Deep ensembles \citep{lakshminarayanan2017simple} provide uncertainty estimates via model disagreement. Predictive uncertainty has also been used to drive exploration in large-scale content personalization \citep{chinta2025variance}. In contrast to these approaches, FARE uses upstream predictor uncertainty as an \textit{execution signal}---not for exploration, and not as a post-hoc fairness constraint.

\textbf{Fair Exposure and Re-Ranking in Recommendation and Retrieval.}

\citet{singh2018fairness} and \citet{biega2018equity} enforce exposure fairness post-hoc; \citet{mehrotra2018towards} counterfactually evaluate recommendation policies that trade relevance against supplier fairness, measured as the spread of recommended artists across popularity bins, and estimate the resulting cost to user satisfaction; \citet{zehlike2017fair}, \citet{xu2023pmmf} and \citet{naghiaei2022cpfair} enforce group-representation, provider max-min and consumer--producer fairness constraints at re-ranking time; and \citet{jaenich2024fairness} act earlier in the pipeline, using corpus-graph adaptive re-ranking to bring documents from under-represented groups into the candidate set, since fair exposure is infeasible when first-stage retrieval returns too few of them. Several of these solve a per-query objective, but not all: \citet{biega2018equity} amortise attention across a sequence of rankings against cumulative relevance, and P-MMF \citep{xu2023pmmf} tracks accumulated provider exposure over a known horizon through an online dual update, against externally specified provider weights. Cumulative allocation state is thus not itself new; FairCo, discussed below, uses it too. P-MMF, however, maximises the exposure of the worst-off provider under a hard per-provider ceiling, excluding providers once their allocation is exhausted. FARE instead applies a signed, graduated correction toward a two-sided target, and none of these methods incorporates the upstream model's prediction uncertainty. FARE's distinction lies in using $\sigma_i$ to scale that correction, without retraining the upstream model.

\textbf{Budget-Constrained Online Allocation.} A parallel line of work in display advertising allocates arriving impressions to budget-constrained advertisers under a hard per-advertiser cap. \citet{shamsi2014online} pose this as convex risk minimisation and derive per-impression dual-price updates in closed form, showing on AOL logs that an exponential update---which raises an advertiser's shadow price when its spend outruns the fraction of the horizon elapsed---dominates greedy allocation on both revenue and smoothness of delivery. The mechanism is close to FARE-PC: both track cumulative consumption against a target over a known horizon and apply a state-dependent correction to the per-arrival score, and both trade constraint satisfaction against immediate value through a single gain. Two things differ. Their constraint is a budget ceiling on a divisible resource, so allocation is a matching problem in which an advertiser can simply be excluded once exhausted; ours is a two-sided exposure target on a fixed slate, where every group is always in contention and the correction must be signed---boosting groups behind target and penalising those ahead. And their correction is scaled by remaining budget alone, whereas FARE-PC scales by the upstream model's predictive uncertainty, so that adjustments concentrate where they cost least. P-MMF \citep{xu2023pmmf}, which builds on the regularised online-allocation framework of \citet{balseiro2021regularized}, applies the same machinery to provider-fair recommendation and differs from FARE-PC in both respects.

\textbf{Control-Theoretic Enforcement and Uncertainty for Fairness.} Two further works are closest in mechanism. FairCo \citep{morik2020controlling} enforces amortised merit-based exposure in dynamic learning-to-rank through an error term proportional to accumulated exposure deficit. FARE-PC shares this proportional-control skeleton, which we make explicit rather than claim as novel; it differs in scaling that deficit by the upstream model's predictive uncertainty $\sigma_i$ and a time-remaining term, against an exogenous rather than relevance-proportional target. PUFR \citep{heuss2023predictive} is, to our knowledge, the first to use predictive uncertainty to bound how far a document may be moved for bias mitigation---the same motivating principle, which we do not claim as new, and with a coefficient likewise multiplying $\sigma$. It bounds each score within $\mu_{q,i} \pm \alpha\sigma_{q,i}$ per query on a static retrieval result, carrying no state between queries; FARE instead uses $\sigma$ as the \emph{gain} of a closed loop whose state is cumulative exposure deficit over a horizon. The uncertainty gain is our claimed contribution; empirical comparison against both remains outstanding (Appendix~\ref{sec:limitations}).

\textbf{Evaluation of Fairness in Recommendation.}
Recent work evaluates fairness with several consumer- and provider-side metrics \citep{malitesta2025fair, diaz2020evaluating}, and public datasets with logged impressions---ContentWise Impressions \citep{perezmaurera2020contentwise} and EB-NeRD \citep{kruse2024ebnerd}---now record what was actually shown, although only ContentWise retains the display order that position-weighted exposure requires. We do not draw on either, and report constraint-violation metrics rather than the consumer- and provider-side fairness metrics used by \citet{malitesta2025fair}.

\section{Conclusion}

We introduced FARE, an uncertainty-aware execution layer that enforces Share-of-Voice constraints on the output of a frozen CTR model. The framing is a reinforcement learning one, adapting the separation of signal generation from constrained execution that is standard in algorithmic trading: the CTR model forecasts, and FARE works an exposure budget over a horizon subject to a target, much as a VWAP executor works an order against a volume schedule. Because FARE acts downstream of the predictor and never enters its training objective, the CTR model can be improved or replaced without retraining the fairness layer, and the layer itself is agnostic to whether the unit of allocation is an item or a group.

The three variants share the pipeline of Eq.~\ref{eq:pipeline} and differ only in how the mean adjustment is computed: FARE-PC by an uncertainty-scaled control law, FARE-ES and FARE-PPO by learned policies. The choice between them depends on whether one or several objectives must be balanced (Sections~\ref{sec:perf_comparison} and~\ref{sec:ablation}). Appendices~\ref{sec:limitations}--\ref{sec:future} state the limitations, the settings to which the results transfer, and the next steps that follow from them.

\section{Disclaimer}

JPMorganChase makes no representation and warranty whatsoever and disclaims all liability, for the completeness, accuracy or reliability of the information contained herein. Any views or opinions expressed herein are solely those of the authors. This document is not intended as investment research or investment advice, or a recommendation, offer or solicitation for the purchase or sale of any security, financial instrument, financial product or service, or to be used in any way for evaluating the merits of participating in any transaction, and shall not constitute a solicitation under any jurisdiction or to any person, if such solicitation under such jurisdiction or to such person would be unlawful.

\clearpage
\bibliographystyle{iclr2026_conference}
\bibliography{references}

\appendix
\setcounter{figure}{0}
\setcounter{table}{0}
\renewcommand{\thefigure}{A\arabic{figure}}
\renewcommand{\thetable}{A\arabic{table}}

\section{Algorithm Details}
\label{app:algorithms}

The CTR-Only baseline follows the deployed ranking procedure of \citet{chinta2025variance}: a multi-head network outputs $(\mu_i, \sigma_i^2)$ for each tile, a score is sampled from each tile's Gaussian, and tiles are ranked by sampled score. We use the variance as predicted, without the exploration multiplier applied there. The FARE variants change this procedure only by adjusting $\mu_i$ before sampling (Eq.~\ref{eq:pipeline}).

\begin{algorithm}[H]
\caption{Direct $(\mu, \sigma^2)$ Parameter Generation}
\label{alg:param_gen}
\small
\begin{algorithmic}[1]
\REQUIRE Number of tiles $K$, users $N$, days $T$; variation scales $\sigma_{\text{day}}, \sigma_{\text{user}}, \sigma_{\text{interaction}}$
\REQUIRE Desired ordering: Tile 1 $>$ Tile 2 $>$ ... $>$ Tile $K$ in Position 1
\STATE \textbf{Set base parameters with ordering constraint:}
\STATE $\bar{\mu} \leftarrow [0.75, 0.65, 0.55, 0.45, 0.35]$ \hfill // Decreasing order ensures tile ordering
\STATE $\bar{\sigma}^2 \leftarrow [0.35, 0.40, 0.50, 0.45, 0.40]$ \hfill // Base variances (can vary independently)
\STATE \textbf{Generate day-level variations:}
\STATE $\delta \sim \mathcal{N}(0, \sigma_{\text{day}}^2)$ with shape $(T, K)$
\STATE \textbf{Generate user-level variations:}
\STATE $\epsilon \sim \mathcal{N}(0, \sigma_{\text{user}}^2)$ with shape $(N, K)$
\STATE $\nu \sim \text{Uniform}(-0.1, 0.1)$ with shape $(N, K)$ \hfill // For variance
\STATE \textbf{Generate user-day interaction:}
\STATE $\eta \sim \mathcal{N}(0, \sigma_{\text{interaction}}^2)$ with shape $(T, N, K)$
\STATE \textbf{Compute final parameters:}
\FOR{each day $d$, user $u$, tile $i$}
    \STATE $\mu_{d,u,i} \leftarrow \text{clip}(\bar{\mu}_i + \delta_{d,i} + \epsilon_{u,i} + \eta_{d,u,i}, 0, 1)$
    \STATE $\sigma^2_{d,u,i} \leftarrow \text{clip}(\bar{\sigma}^2_i + \nu_{u,i}, 0.2, 0.8)$
\ENDFOR
\RETURN $\{(\mu_{d,u,i}, \sigma^2_{d,u,i})\}$
\end{algorithmic}
\end{algorithm}

\begin{algorithm}[H]
\caption{FARE-PC: Uncertainty-Weighted Proportional Control}
\label{alg:vwap}
\small
\begin{algorithmic}[1]
\REQUIRE CTR scores $(\mu, \sigma)$, target SOV $\tau$, gain $\alpha$, horizon $T_{\text{total}}$
\STATE Initialize $\text{counts} \leftarrow \mathbf{0}$, $t \leftarrow 0$
\FOR{each impression}
    \STATE $\text{SOV} \leftarrow \text{counts} / \max(t, 1)$ \hfill // Current SOV
    \STATE $\text{deficit} \leftarrow \tau - \text{SOV}$ \hfill // Positive if behind target
    \STATE $\text{remaining} \leftarrow T_{\text{total}} - t$ \hfill // Time left
    \STATE $\text{adj} \leftarrow \alpha \cdot \text{deficit} \cdot \sigma / \text{remaining}$ \hfill // Uncertainty-weighted
    \STATE $\tilde{\mu} \leftarrow \mu + \text{adj}$ \hfill // Additive adjustment
    \STATE $\tilde{s}_i$ $\sim \mathcal{N}(\tilde{\mu}_i, \sigma_i^2)$ for all $i$ \hfill // \textbf{Sample from adjusted distribution}
    \STATE $\text{winner} \leftarrow$ $\arg\max(\tilde{s})$ \hfill // Rank by sampled scores
    \STATE $\text{counts}[\text{winner}] \leftarrow \text{counts}[\text{winner}] + 1$
    \STATE $t \leftarrow t + 1$
\ENDFOR
\end{algorithmic}
\end{algorithm}

\begin{algorithm}[H]
\caption{FARE-ES: Evolution Strategy Policy}
\label{alg:fare_es}
\small
\begin{algorithmic}[1]
\REQUIRE CTR scores $(\mu, \sigma^2)$, target SOV $\tau$, population size $P$, noise std $\sigma_{\text{ES}}$, gain $\alpha$
\STATE Initialize policy network weights $\theta$
\FOR{generation $= 1$ to $G$}
    \FOR{$p = 1$ to $P$}
        \STATE Sample perturbation: $\epsilon_p \sim \mathcal{N}(0, I)$
        \STATE $\theta_p \leftarrow \theta + \sigma_{\text{ES}} \cdot \epsilon_p$
        \STATE Evaluate fitness $F_p$ by running episode with $\pi_{\theta_p}$:
        \STATE \quad Build state $X$ $= [\mu, \sigma^2, \text{SOV}, \Delta\text{SOV},$ $(T-t)/T$$]$
        \STATE \quad Get action $a = \pi_{\theta_p}(X)$
        \STATE \quad Apply: $\tilde{\mu} \leftarrow \mu + \alpha \cdot a$ \hfill // Additive adjustment
        \STATE \quad Sample $\tilde{s}_i \sim \mathcal{N}(\tilde{\mu}_i, \sigma_i^2)$ for all $i$; rank by $\tilde{s}$, compute reward
    \ENDFOR
    \STATE Update: $\theta \leftarrow \theta + \frac{\eta}{P \sigma_{\text{ES}}} \sum_{p=1}^{P} F_p \cdot \epsilon_p$
\ENDFOR
\RETURN Policy $\pi_\theta$
\end{algorithmic}
\end{algorithm}

\begin{algorithm}[H]
\caption{FARE-PPO: Proximal Policy Optimization}
\label{alg:fare}
\small
\begin{algorithmic}[1]
\REQUIRE CTR scores $(\mu, \sigma^2)$, target SOV $\tau$, gain $\alpha$, learning rate $\eta$, clip $\epsilon$, PPO epochs $E$
\STATE Initialize actor-critic network $\pi_\theta$ with shared features
\FOR{episode $= 1$ to $M$}
    \STATE Reset SOV counters; collect trajectory $\mathcal{T} = \{(X_t, a_t, r_t, \log\pi_\theta(a_t|X_t))\}$
    \FOR{each impression $(d, u)$}
        \STATE $X$ $\leftarrow [\mu, \sigma^2, \text{SOV}, \Delta\text{SOV}, (T-t)/T]$
        \STATE $a \sim \pi_\theta(X)$; $\tilde{\mu} \leftarrow \mu + \alpha \cdot a$
        \STATE Sample $\tilde{s}_i \sim \mathcal{N}(\tilde{\mu}_i, \sigma_i^2)$ for all $i$; rank tiles by $\tilde{s}$, observe reward $r$, next state $X'$
    \ENDFOR
    \STATE Compute advantages $\hat{A}_t$ using GAE($\lambda$)
    \FOR{PPO epoch $= 1$ to $E$}
        \STATE $r_t(\theta) = \frac{\pi_\theta(a_t|X_t)}{\pi_{\theta_{\text{old}}}(a_t|X_t)}$ \hfill // Probability ratio
        \STATE $L^{\text{CLIP}} = \min(r_t \hat{A}_t, \text{clip}(r_t, 1{-}\epsilon, 1{+}\epsilon) \hat{A}_t)$
        \STATE Update $\theta$ via $\nabla_\theta (L^{\text{CLIP}} + c_1 L^{\text{VF}} - c_2 H[\pi_\theta])$
    \ENDFOR
\ENDFOR
\end{algorithmic}
\end{algorithm}

\section{Input Generation Details}
\label{app:inputs}

\subsection{Synthetic Generator Constants}
\label{app:synth_constants}
In Eq.~\ref{eq:synth}, the base means are $\bar{\mu}_i \in [0.75, 0.65, 0.55, 0.45, 0.35]$, set in decreasing order to enforce the tile ordering, and the base variances are $\bar{\sigma}^2_i \in [0.35, 0.40, 0.50, 0.45, 0.40]$, which vary independently of the means. Day-level noise $\delta \sim \mathcal{N}(0, \sigma^2_{\text{day}})$, user-level noise $\epsilon \sim \mathcal{N}(0, \sigma^2_{\text{user}})$, and user--day interaction noise $\eta \sim \mathcal{N}(0, \sigma^2_{\text{intr.}})$ are added with $\sigma_{\text{day}} = 0.03$, $\sigma_{\text{user}} = 0.08$, and $\sigma_{\text{intr.}} = 0.02$. All values are clipped to $\mu \in [0,1]$ and $\sigma^2 \in [0.2, 0.8]$.

\begin{figure}[H]
\centering
\vspace{-4pt}
\includegraphics[width=0.82\textwidth]{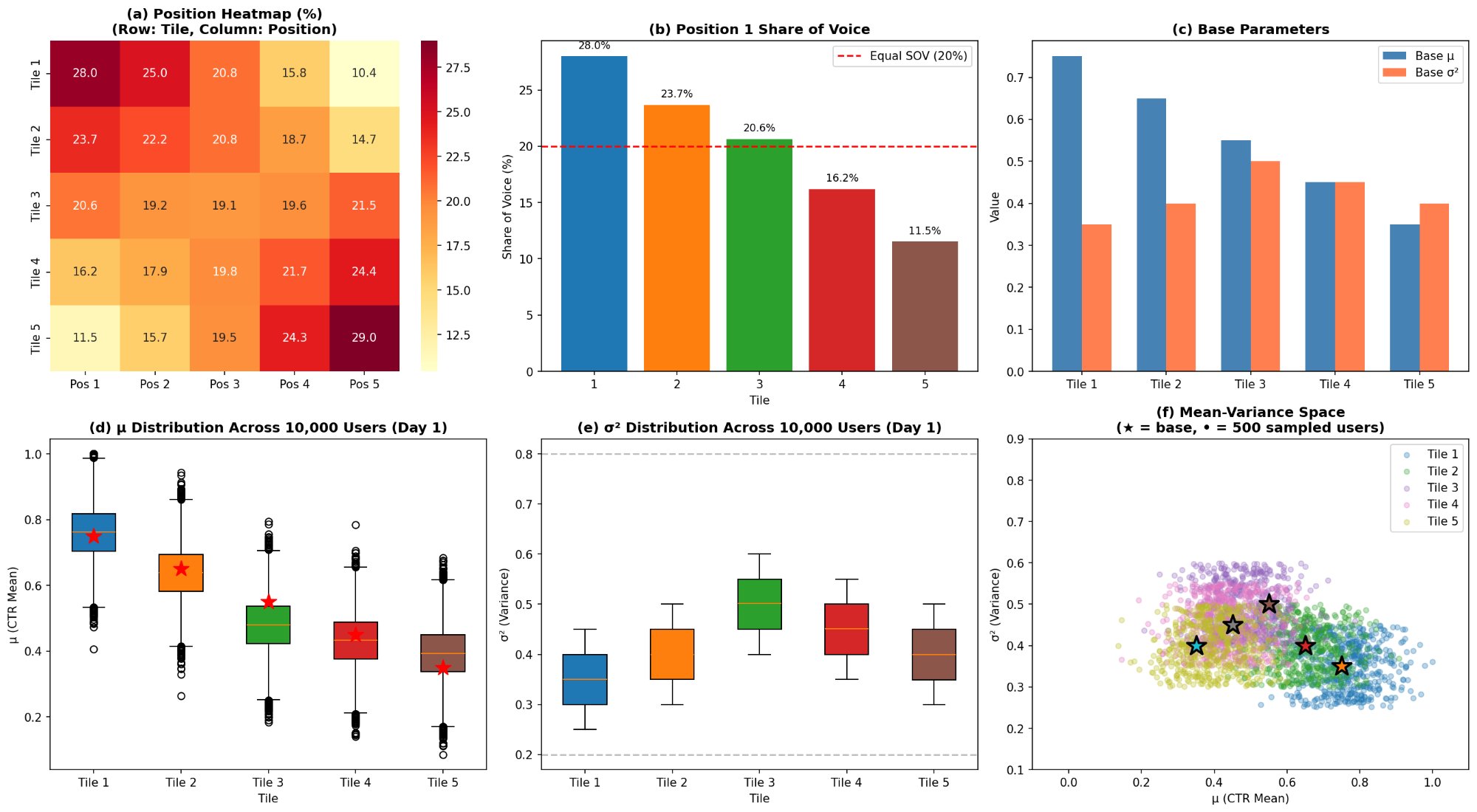}
\caption{Synthetic CTR estimates $(\mu, \sigma^2)$ and resulting SOV imbalance. \textbf{Top row:} (a) Position heatmap. (b) Position 1 SOV distribution. (c) Base $\mu$ and $\sigma^2$ per tile. \textbf{Bottom row:} (d--f) User-level variation in $\mu$, $\sigma^2$, and mean-variance space across 10,000 users.}
\label{fig:param_variation}
\vspace{-8pt}
\end{figure}

\subsection{KuaiRand-Pure Tile Construction}
\label{app:tile_construction}
We use KuaiRand-Pure's randomly exposed impressions, in which videos are sampled uniformly from the candidate pool, so the predictor's CTR estimates are not biased by the production recommender's exposure policy. Tile $i$ is the $i$-th most frequent Level-1 category by total interaction count in the training period; a video is assigned to exactly one Level-1 category via the first entry of its tag metadata, and videos whose primary category falls outside the top~5 are dropped rather than pooled into an ``other'' tile. The target for triple $(u,d,i)$ is $\text{clicks}_{u,d,i} / \text{impressions}_{u,d,i}$, defined only where the denominator is non-zero; the remaining cells are sentinel-marked and excluded from the loss by the mask $\mathcal{M}$ in Eq.~\ref{eq:gnll}. This aggregation is a modelling choice that matches the group-exposure setting of Section~\ref{sec:setting}---the exposure constraint is defined over categories, so the predictor is defined over categories---and we note in Appendix~\ref{sec:discussion} what it costs.

\begin{figure}[H]
\centering
\vspace{-4pt}
\includegraphics[width=0.82\textwidth]{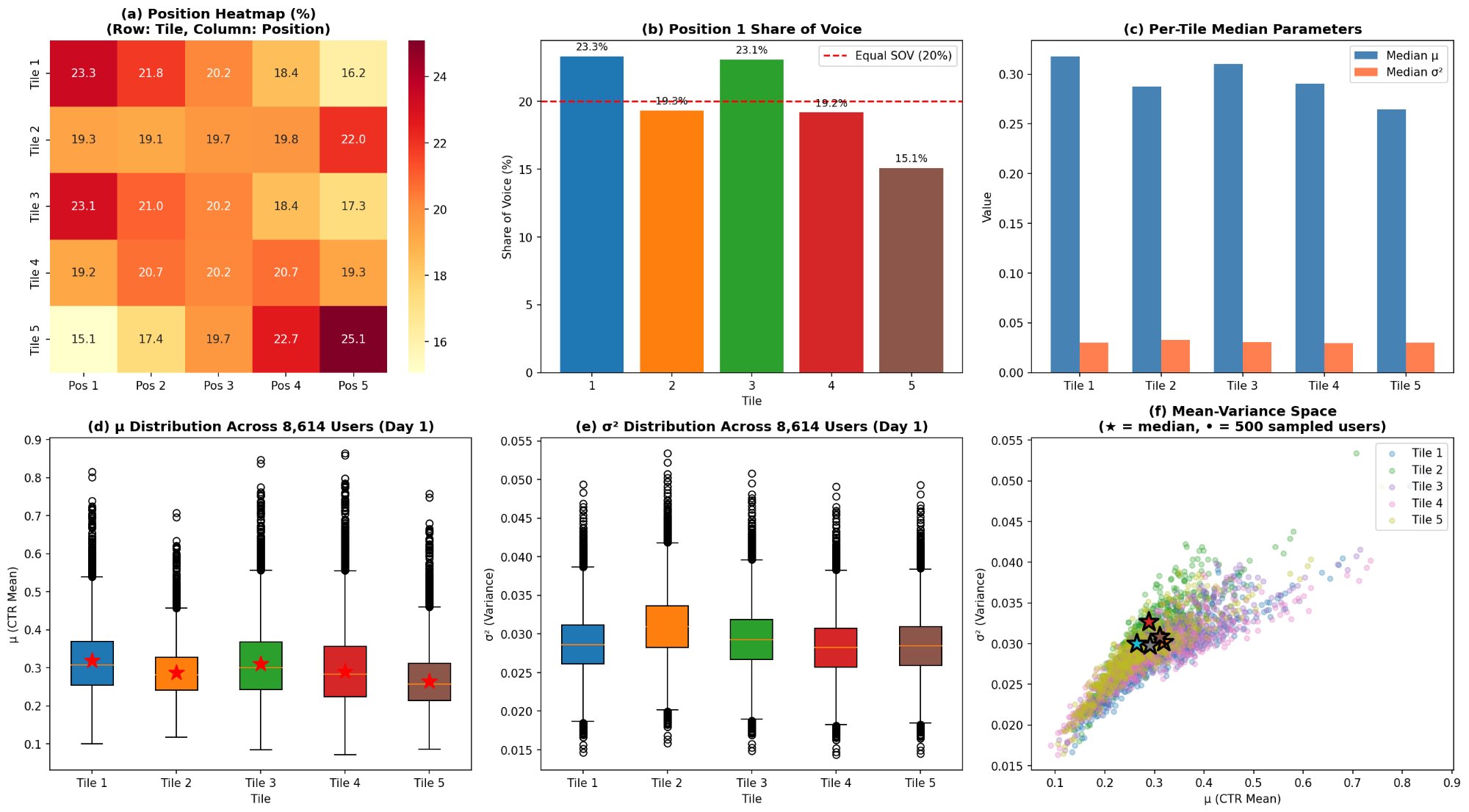}
\caption{KuaiRand-Pure MLP predictions. \textbf{Top row:} (a)~Position heatmap shows non-monotonic ordering (T1 $\approx$ T3 $>$ T2 $\approx$ T4 $>$ T5). (b)~Position~1 SOV reveals a milder fairness gap than synthetic data. (c)~Per-tile median $\mu$ and $\sigma^2$. \textbf{Bottom row:} (d)~$\mu$ distribution across 8,614 users (Day~1). (e)~$\sigma^2$ distribution. (f)~Mean-variance space.}
\label{fig:kuairand_params}
\vspace{-8pt}
\end{figure}

\section{Additional Experimental Results}
\label{app:extra}

\subsection{Uniform SOV Targets}
\label{app:uniform_results}

\begin{table}[H]
\centering
\caption{KuaiRand-Pure SOV distribution on test set (\%), one representative seed. Target SOV: 20\% per tile; T1--T5 denote the five category tiles.
}
\label{tab:kuairand_sov}
\begin{tabular}{l|ccccc|c|c}
\toprule
 & \textbf{T1} & \textbf{T2} & \textbf{T3} & \textbf{T4} & \textbf{T5} & \textbf{SOV Error} & \textbf{Reduction} \\
\midrule
CTR-Only & 23.4 & 19.2 & 22.7 & 19.6 & 15.1 & 0.122 & -- \\
FARE-PC & \textbf{20.3} & \textbf{19.9} & \textbf{20.2} & \textbf{19.9} & \textbf{19.7} & \textbf{0.010} & \textbf{91.8\%} \\
FARE-ES & 22.1 & 18.9 & 22.7 & 19.6 & 16.7 & 0.099 & 18.9\% \\
FARE-PPO & 21.8 & 19.2 & 21.5 & 19.9 & 17.6 & 0.089 & 27.0\% \\
\midrule
Target & 20.0 & 20.0 & 20.0 & 20.0 & 20.0 & 0.000 & -- \\
\bottomrule
\end{tabular}
\end{table}

\subsection{Non-Uniform SOV Targets}
\label{app:nonuniform_results}

\begin{table}[H]
\centering
\caption{Non-uniform SOV targets (synthetic): test set evaluation metrics (mean $\pm$ std). Targets: $\tau = [10, 15, 20, 25, 30]\%$.}
\label{tab:nonuniform_metrics}
\begin{tabular}{l|c|c|c|c|c}
\toprule
\textbf{Method} & \textbf{SOV Error} $\downarrow$ & \textbf{PWCL\%} $\downarrow$ & \textbf{Pos.\ Disp.} & \textbf{Kendall $\tau$} & \textbf{Top-1 Chg} \\
\midrule
CTR-Only & 0.547 $\pm$ 0.003 & 0.0\% & 0.00 & 0.00 & 0.0\% \\
FARE-PC & \textbf{0.116 $\pm$ 0.001} & 5.31 $\pm$ 0.02\% & 1.62 $\pm$ 0.001 & 0.508 $\pm$ 0.000 & 80.8 $\pm$ 0.1\% \\
FARE-ES & 0.120 $\pm$ 0.003 & 4.92 $\pm$ 0.18\% & 1.61 $\pm$ 0.003 & 0.505 $\pm$ 0.001 & 80.6 $\pm$ 0.1\% \\
FARE-PPO & 0.367 $\pm$ 0.144 & 2.24 $\pm$ 1.59\% & 1.57 $\pm$ 0.023 & 0.490 $\pm$ 0.009 & 79.0 $\pm$ 0.9\% \\
\bottomrule
\end{tabular}
\end{table}

\begin{table}[H]
\centering
\caption{Non-uniform SOV targets (synthetic): final SOV distribution on test set (\%), one representative seed. Targets: $\tau = [10, 15, 20, 25, 30]\%$.}
\label{tab:nonuniform_sov}
\begin{tabular}{l|ccccc|c|c}
\toprule
 & \textbf{T1} & \textbf{T2} & \textbf{T3} & \textbf{T4} & \textbf{T5} & \textbf{SOV Error} & \textbf{Reduction} \\
\midrule
CTR-Only & 28.0 & 23.6 & 20.7 & 16.1 & 11.5 & 0.547 & -- \\
FARE-PC & 14.6 & 16.2 & 19.0 & 22.8 & 27.4 & \textbf{0.116} & \textbf{78.8\%} \\
FARE-ES & 16.0 & 14.1 & 19.2 & 24.0 & 26.7 & 0.120 & 78.1\% \\
FARE-PPO & 25.6 & 16.2 & 19.1 & 20.7 & 18.5 & 0.367 & 32.9\% \\
\midrule
Target & 10.0 & 15.0 & 20.0 & 25.0 & 30.0 & 0.000 & -- \\
\bottomrule
\end{tabular}
\end{table}

\subsection{Multi-Position SOV Weighting}
\label{app:multipos_results}

\begin{table}[H]
\centering
\caption{Multi-position SOV: performance across weighting schemes (uniform target $\tau{=}20\%$). Mean $\pm$ std over seeds.
}
\label{tab:multipos_sov}
\resizebox{\textwidth}{!}{%
\begin{tabular}{l|c|ccc|ccc|ccc}
\toprule
& & \multicolumn{3}{c|}{\textbf{FARE-PC}} & \multicolumn{3}{c|}{\textbf{FARE-ES}} & \multicolumn{3}{c}{\textbf{FARE-PPO}} \\
\textbf{Scheme} & \textbf{Base} & \textbf{Err.} & \textbf{Red.} & \textbf{PWCL} & \textbf{Err.} & \textbf{Red.} & \textbf{PWCL} & \textbf{Err.} & \textbf{Red.} & \textbf{PWCL} \\
\midrule
Top-1 & .247 & .024\tiny{$\pm$.000} & \textbf{90.3\%} & 2.64\% & .034\tiny{$\pm$.004} & 86.2\% & 2.79\% & .155\tiny{$\pm$.080} & 37.2\% & 1.50\% \\
Top-3 & .137 & .040\tiny{$\pm$.001} & 70.8\% & 2.27\% & .036\tiny{$\pm$.005} & \textbf{73.7\%} & 2.39\% & .114\tiny{$\pm$.023} & 16.8\% & 0.63\% \\
DCG & .061 & .032\tiny{$\pm$.000} & \textbf{47.5\%} & 1.46\% & .067\tiny{$\pm$.005} & $-$9.8\% & $-$0.32\% & .063\tiny{$\pm$.001} & $-$3.3\% & $-$0.08\% \\
\bottomrule
\end{tabular}%
}
\end{table}

\begin{table}[H]
\centering
\caption{KuaiRand-Pure multi-pos SOV: perf.\ across wtg.\ schemes (unif.\ target $\tau{=}20\%$). Mean $\pm$ std.
}
\label{tab:kuairand_multipos}
\resizebox{\textwidth}{!}{%
\begin{tabular}{l|c|ccc|ccc|ccc}
\toprule
& & \multicolumn{3}{c|}{\textbf{FARE-PC}} & \multicolumn{3}{c|}{\textbf{FARE-ES}} & \multicolumn{3}{c}{\textbf{FARE-PPO}} \\
\textbf{Scheme} & \textbf{Base} & \textbf{Err.} & \textbf{Red.} & \textbf{PWCL} & \textbf{Err.} & \textbf{Red.} & \textbf{PWCL} & \textbf{Err.} & \textbf{Red.} & \textbf{PWCL} \\
\midrule
Top-1 & .122 & .010\tiny{$\pm$.001} & \textbf{91.8\%} & 0.19\% & .099\tiny{$\pm$.009} & 18.9\% & 0.06\% & .089\tiny{$\pm$.032} & 27.0\% & 0.10\% \\
Top-3 & .063 & .017\tiny{$\pm$.001} & \textbf{72.6\%} & 0.16\% & .055\tiny{$\pm$.012} & 12.7\% & 0.04\% & .047\tiny{$\pm$.016} & 25.4\% & 0.09\% \\
DCG & .030 & .015\tiny{$\pm$.001} & \textbf{49.6\%} & 0.11\% & .029\tiny{$\pm$.001} & 3.3\% & 0.01\% & .025\tiny{$\pm$.009} & 16.7\% & 0.05\% \\
\bottomrule
\end{tabular}%
}
\end{table}

\subsection{Multi-Objective: SOV and Diversity}
\label{app:multiobj_results}

\begin{table}[H]
\centering
\caption{Multi-objective $\lambda_{\text{div}}$ sweep: SOV Error, PWCL, and Category Diversity across methods. CTR-Only baseline: SOV Err = 0.247, Diversity = 2.300.}
\label{tab:multiobj}
\begin{tabular}{c|ccc|ccc|ccc}
\toprule
& \multicolumn{3}{c|}{\textbf{FARE-PC}} & \multicolumn{3}{c|}{\textbf{FARE-ES}} & \multicolumn{3}{c}{\textbf{FARE-PPO}} \\
$\boldsymbol{\lambda_{\textbf{div}}}$ & \textbf{SOV} $\downarrow$ & \textbf{PWCL} & \textbf{Div} $\uparrow$ & \textbf{SOV} $\downarrow$ & \textbf{PWCL} & \textbf{Div} $\uparrow$ & \textbf{SOV} $\downarrow$ & \textbf{PWCL} & \textbf{Div} $\uparrow$ \\
\midrule
0.0 & \textbf{.024} & 2.64\% & 2.383 & .034 & 2.71\% & 2.398 & .172 & 1.34\% & 2.339 \\
0.5 & \textbf{.010} & 2.83\% & 2.402 & .048 & 2.85\% & 2.409 & .099 & 2.41\% & 2.414 \\
1.0 & .043 & 3.03\% & 2.405 & .049 & 2.83\% & 2.409 & .122 & 2.28\% & 2.416 \\
2.0 & .175 & 3.87\% & 2.427 & .063 & 2.86\% & 2.415 & .120 & 1.86\% & 2.399 \\
5.0 & .632 & 6.69\% & 2.471 & .088 & 3.10\% & 2.427 & .119 & 2.19\% & 2.414 \\
\bottomrule
\end{tabular}%

\end{table}

\subsection{Effect of Increasing the Number of Tiles}
\label{app:K_results}

\begin{table}[H]
\centering
\caption{Performance as $K$ (number of tiles) increases. Uniform targets ($\tau = 1/K$). Mean $\pm$ std over seeds.
}
\label{tab:scalability_K}
\resizebox{\textwidth}{!}{%
\begin{tabular}{c|c|ccc|ccc|ccc}
\toprule
& & \multicolumn{3}{c|}{\textbf{FARE-PC}} & \multicolumn{3}{c|}{\textbf{FARE-ES}} & \multicolumn{3}{c}{\textbf{FARE-PPO}} \\
\textbf{K} & \textbf{Base SOV Err.} & \textbf{Err.} $\downarrow$ & \textbf{Red.} & \textbf{PWCL} & \textbf{Err.} $\downarrow$ & \textbf{Red.} & \textbf{PWCL} & \textbf{Err.} $\downarrow$ & \textbf{Red.} & \textbf{PWCL} \\
\midrule
3 & 0.252 & .014\tiny{$\pm$.000} & \textbf{94.4\%} & 2.45\% & .030\tiny{$\pm$.012} & 88.1\% & 2.41\% & .169\tiny{$\pm$.086} & 32.9\% & 0.96\% \\
5 & 0.247 & .024\tiny{$\pm$.000} & \textbf{90.3\%} & 2.64\% & .034\tiny{$\pm$.004} & 86.2\% & 2.79\% & .155\tiny{$\pm$.080} & 37.2\% & 1.50\% \\
7 & 0.249 & .039\tiny{$\pm$.001} & \textbf{84.3\%} & 2.73\% & .055\tiny{$\pm$.003} & 77.9\% & 2.78\% & .158\tiny{$\pm$.047} & 36.5\% & 1.82\% \\
10 & 0.253 & .058\tiny{$\pm$.001} & \textbf{77.1\%} & 2.87\% & .066\tiny{$\pm$.026} & 73.9\% & 3.49\% & .251\tiny{$\pm$.012} & 0.8\% & $-$0.11\% \\
\bottomrule
\end{tabular}%
}
\end{table}

\subsection{Control-Gain Sensitivity}
\label{app:alpha_results}

\begin{figure}[H]
\centering
\vspace{-4pt}
\includegraphics[width=\textwidth]{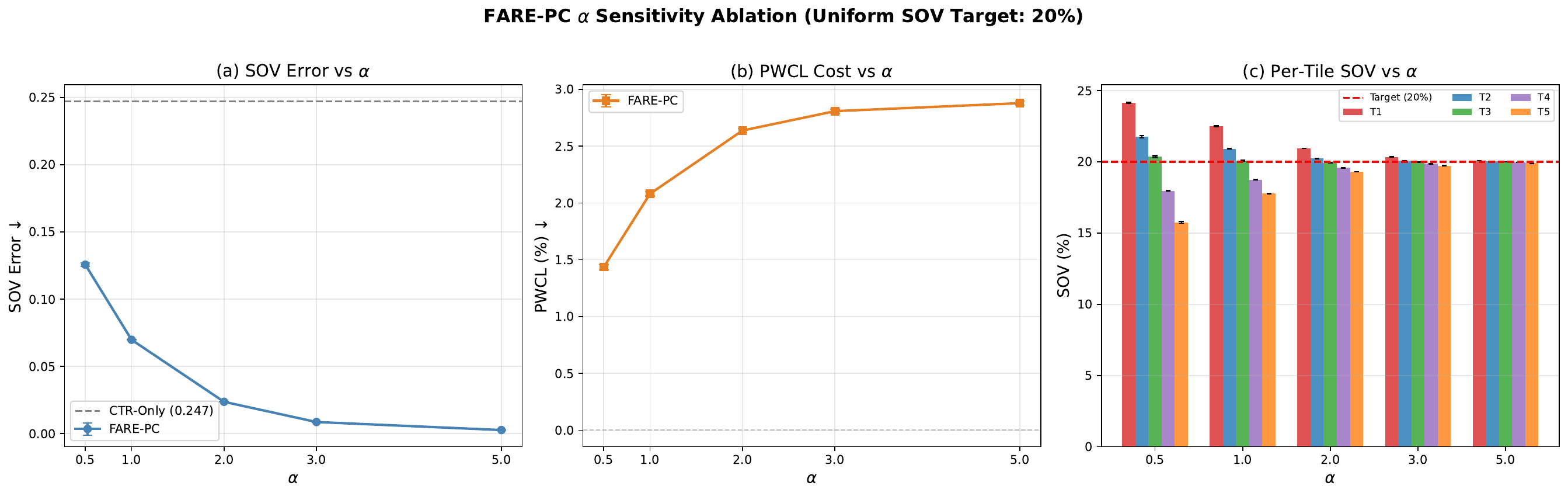}
\caption{FARE-PC $\alpha$ sensitivity under \textbf{uniform} SOV targets (20\%). (a)~SOV Error decreases monotonically with $\alpha$. (b)~PWCL cost increases with diminishing marginal cost. (c)~Per-tile SOV converges to target. Dashed line: CTR-Only baseline error (0.247).}
\label{fig:alpha_uniform}
\vspace{-8pt}
\end{figure}

\subsection{Uncertainty Miscalibration}
\label{app:sigma_results}

\begin{figure}[H]
\centering
\begin{minipage}[c]{0.38\textwidth}
\includegraphics[width=\linewidth]{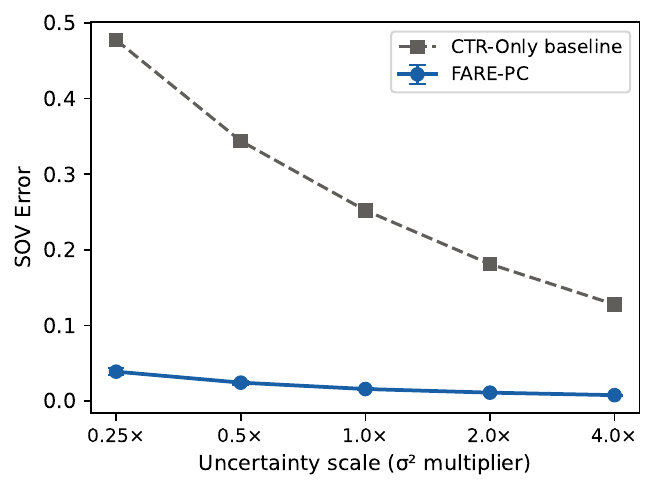}
\end{minipage}
\hfill
\begin{minipage}[c]{0.58\textwidth}
\caption{FARE-PC SOV Error vs CTR-Only across uncertainty scale factors $s \in \{0.25\times\text{--}4\times\}$.}
\label{fig:sigma_sensitivity}
\end{minipage}
\end{figure}

\section{Limitations}
\label{sec:limitations}

We state the limitations of this study explicitly, since they bound how far the present results can be read.

\textbf{No comparison against published fair re-ranking systems.}
Our baselines (CTR-Only, Max-Deficit, Quota-CTR) are reference policies we implemented to span the trade-off, not competitive methods from the literature. We do not benchmark against P-MMF \citep{xu2023pmmf}, CPFair \citep{naghiaei2022cpfair}, fairness-aware adaptive re-ranking \citep{jaenich2024fairness}, FA*IR \citep{zehlike2017fair}, FairCo \citep{morik2020controlling}, or PUFR \citep{heuss2023predictive}. These methods differ in how directly they transfer. P-MMF, CPFair, FA*IR and fairness-aware adaptive re-ranking select a slate from a larger candidate pool and target parity, a minimum proportion, or externally weighted exposure, so porting them to a fixed $K$-tile surface requires design decisions that would themselves need justification. FairCo, PUFR and \citet{singh2018fairness}, which we also do not benchmark, instead re-order a fixed set and so apply directly. FairCo defines fairness relative to estimated merit, but substituting the target $\tau_i$ for merit yields an exogenous-target controller on a fixed slate, which makes it the most direct comparison and the most significant omission. PUFR addresses a two-group setting without an exposure target, and \citeauthor{singh2018fairness} enforce their constraint per impression in expectation rather than over the horizon; Quota-CTR, by contrast, enforces the target as a hard cumulative quota. Until these comparisons exist, our results should be read as characterising the fairness--engagement frontier reachable by graduated uncertainty-weighted control relative to naive enforcement, not as evidence of superiority over the state of the art.

\textbf{Non-standard metrics.}
\label{app:metric_relations}
SOV Error and PWCL are constraint-violation and utility-loss measures defined for our setting (Section~\ref{sec:metrics}). We report neither standard accuracy metrics (NDCG@$k$, Recall@$k$) nor the fairness metrics used elsewhere in the community (consumer- and provider-side utility and exposure gaps \citep{malitesta2025fair}, provider max-min utility \citep{xu2023pmmf}, expected exposure \citep{diaz2020evaluating}). This limits the comparability of our numbers against published results.

SOV Error, $\sum_i |\text{SOV}_i - \tau_i|$, is the $L_1$ distance between realised and target exposure over groups, i.e.\ twice their total variation distance. It aggregates the per-group violations, whereas Eq.~\ref{eq:constrained} constrains each group; the per-tile distributions (Tables~\ref{tab:sov_distribution}, \ref{tab:kuairand_sov} and \ref{tab:nonuniform_sov}) give the per-group check. Under the DCG weighting of Eq.~\ref{eq:multipos_sov}, exposure is position-weighted as in the fair-ranking literature \citep{singh2018fairness, diaz2020evaluating}, and SOV Error, like the expected-exposure loss of \citet{diaz2020evaluating}, measures a distance between achieved and target exposure; it differs in using $L_1$ distance over groups rather than squared $L_2$ distance over documents, and an exogenous rather than relevance-derived target. PWCL is the DCG-weighted difference between the sampled scores of the CTR-Only ordering and those of the re-ranked ordering, i.e.\ a DCG loss against the unconstrained sampled ranking rather than the $\mu$-sorted, CTR-optimal one. Because the two policies draw their scores independently, PWCL is noisy and can fall slightly below zero when a policy barely adjusts (Tables~\ref{tab:multipos_sov} and \ref{tab:scalability_K}). It is a model-based estimate of utility loss rather than a measured one, since no counterfactual click data exists for the rankings FARE produces.

\textbf{Policy-gradient learning is unreliable here.}
\label{app:ppo_failure}
FARE-PPO trails FARE-ES on synthetic data but leads it on KuaiRand-Pure (Tables~\ref{tab:baselines}, \ref{tab:kuairand_metrics} and \ref{tab:kuairand_multipos}), with high variance across seeds in both. We attribute its instability to weak per-step credit assignment, though we have not ruled out tuning. The reward of Eq.~\ref{eq:reward} is computed at every step, but each action shifts cumulative SOV by at most $1/t$, so any one action's effect on it is small relative to sampling noise and policy-gradient estimates are correspondingly high-variance. FARE-PC sidesteps the issue entirely by reacting to the current deficit rather than learning from a return, and FARE-ES tolerates it on synthetic data because population-based search scores whole trajectories rather than attributing credit to individual steps \citep{salimans2017evolution}. FARE-PPO is the deep RL instantiation of the formulation and the strongest learned policy on KuaiRand-Pure; characterising when policy-gradient training is reliable in fairness-constrained ranking is part of what the formulation makes possible.

\textbf{Offline evaluation and user-side effects.}
All results are offline. We measure engagement loss as predicted CTR degradation, not observed clicks, and we do not evaluate the distribution of that loss across users: an aggregate PWCL of 2.6\% (FARE-PC, synthetic) is consistent with a small subset of users absorbing a much larger relevance penalty. Consumer-side fairness \citep{naghiaei2022cpfair} is therefore not assessed. Results are averaged over three seeds without significance testing.

\section{Discussion}
\label{sec:discussion}

The following properties of the study are consequences of design choices rather than defects, but they determine the settings to which the results transfer.

\textbf{Small, fully-exposed item set.}
All experiments use $K \leq 10$ groups, and every candidate appears in every slate, so the problem is purely one of ordering. Results do not speak to the common recommendation regime in which a catalogue of thousands is filtered to a slate of tens, where selection effects, long-tail coverage, and per-item exposure sparsity dominate, and where achievable fairness is bounded by what first-stage retrieval surfaces \citep{jaenich2024fairness}. The graceful degradation observed from $K=3$ to $K=10$ (Section~\ref{sec:scalability_K}) is evidence about the controller's behaviour as the action space grows, not evidence that the approach transfers to large catalogues.

\textbf{Dataset scope and aggregation.}
We evaluate on one synthetic generator and one real dataset. KuaiRand-Pure is aggregated to the top-5 Level-1 categories, which matches the group-exposure setting of Section~\ref{sec:setting} but discards item-level structure and, with it, any ability to measure long-tail item exposure; videos outside those categories are dropped rather than pooled. KuaiRand-Pure logs impressions, but we use them only to train the predictor; the exposure FARE's re-rankings would receive is simulated from model predictions, as it must be offline for any new ranking policy. Datasets that also log display positions \citep{perezmaurera2020contentwise} would allow position-weighted exposure to be measured for the logged rankings, which we do not attempt. Because all variants are evaluated on, and the learned ones trained against, a frozen predictive model acting as a replay simulator, their reported performance inherits that model's errors: no counterfactual click feedback enters the loop, and estimated engagement loss is model-based rather than measured.

\textbf{One upstream model.}
The modularity claim---that FARE composes with any uncertainty-aware CTR model---is supported architecturally and by the $\sigma^2$ rescaling study (Section~\ref{sec:robustness_sigma}), but we evaluate a single multi-head Gaussian MLP. We do not swap in deep ensembles, MC-dropout, or Bayesian last-layer models, nor do we assess the calibration of the variances the MLP produces; the miscalibration study perturbs $\sigma^2$ synthetically rather than comparing genuinely differently-calibrated models.

\textbf{Choice of experimental settings.}
The non-uniform target vector $\tau=[10,15,20,25,30]\%$ is a deliberately adversarial stress test---it inverts the CTR ordering on synthetic data---rather than a target drawn from a real mandate, and the diversity weights $\lambda_{\text{div}}$ are swept over an arbitrary grid. These settings probe controller behaviour; they should not be read as representative operating points.

\section{Future Work}
\label{sec:future}

The agenda below follows Appendices~\ref{sec:limitations} and~\ref{sec:discussion} in order: the first three items address the limitations, the next four the scoping properties, and the last collects extensions to the method itself.

\textbf{Benchmarking against fair re-ranking baselines.}
The priority is a common-protocol comparison against the methods that apply directly to a fixed slate, FairCo \citep{morik2020controlling} with $\tau_i$ substituted for merit and the per-impression exposure constraints of \citet{singh2018fairness}, followed by ported versions of P-MMF \citep{xu2023pmmf} and CPFair \citep{naghiaei2022cpfair}. Since FairCo shares FARE-PC's controller skeleton, an ablation isolating the uncertainty gain---FARE-PC with $\sigma_i$ replaced by a constant---would directly quantify the contribution we claim. PUFR \citep{heuss2023predictive} ran the analogous ablation for per-query bias mitigation, replacing $\sigma$ with its mean, and found that per-item uncertainty gave the better trade-off.

\textbf{Reporting on community-standard metrics.}
Alongside SOV Error and PWCL, reporting expected exposure \citep{diaz2020evaluating}, provider max-min utility \citep{xu2023pmmf}, the consumer- and provider-side gaps of \citet{malitesta2025fair} and NDCG@$k$ would make the numbers directly comparable
with published re-ranking results, which the present metric choice precludes.

\textbf{Consumer-side and per-user effects.}
Reporting the distribution of PWCL across users, and jointly constraining provider exposure and per-user utility loss, would establish whether SOV enforcement concentrates its cost on a minority of users. Moving from predicted CTR to observed clicks, through production A/B testing, would replace the model-based estimate of engagement loss with a measured one.

\textbf{Catalogues larger than the slate.}
Extending to the regime where the candidate pool exceeds the slate requires defining SOV over a
subset-selection decision rather than a permutation, and raises the long-tail exposure questions
that the current category aggregation forecloses. In that regime fairness also depends on which items are retrieved, not only on how they are ordered, and fairness-aware adaptive re-ranking \citep{jaenich2024fairness}, which acts on candidate discovery and so has no counterpart on a fixed slate, becomes the natural comparison.

\textbf{Impression-logged datasets.}
Evaluating on ContentWise Impressions \citep{perezmaurera2020contentwise} and EB-NeRD \citep{kruse2024ebnerd} would let exposure be measured on logged impressions rather than purely simulated. ContentWise records each list's row position and the order of items within it, so position-weighted SOV of the logged rankings can be computed on it directly, although its impressions are logged at series level and omit rows inserted by the service provider. EB-NeRD shuffles the order of in-view articles, so it supports exposure counts but not position-weighted SOV.

\textbf{Model-agnosticism under test.}
Substituting deep ensembles \citep{lakshminarayanan2017simple}, MC-dropout and efficient alternatives to it \citep{chinta2025variance}, and Bayesian \citep{blundell2015weight} predictors for the Gaussian MLP, with calibration diagnostics on the resulting variances, would convert the modularity claim from an architectural argument into an empirical one.

\textbf{Principled experimental settings.}
Target vectors drawn from real contractual mandates, rather than the adversarial
$\tau$ used here as a stress test, and a $\lambda_{\text{div}}$ range derived from an operator's
tolerance rather than an arbitrary grid, would make the reported operating points representative
rather than merely illustrative.

\textbf{Method extensions.}
FARE-PC computes tile adjustments independently, ignoring competitive ranking dynamics; joint tile optimisation is a natural next step. Further directions include offline RL \citep{levine2020offline}, time-varying targets $\tau_i(t)$, and integral and derivative terms beyond proportional control.

\newpage
\end{document}